\documentclass[lettersize,journal]{IEEEtran}
\usepackage{amsmath,amssymb,amsfonts}
\usepackage{algorithmic}
\usepackage{algorithm}
\usepackage{array}
\usepackage[caption=false,font=normalsize,labelfont=sf,textfont=sf]{subfig}
\usepackage{textcomp}
\usepackage{stfloats}
\usepackage{url}
\usepackage{verbatim}
\usepackage{graphicx}
\usepackage{cite}
\usepackage{booktabs}
\usepackage{multirow}
\usepackage{xcolor}
\usepackage{xspace}
\usepackage[hidelinks]{hyperref}
\usepackage{orcidlink}
\usepackage{marvosym}
\colorlet{orcidlogocol}{black}
\newcommand{\liu}[1]{\textcolor[rgb]{0.90, 0.0, 0.0}{#1}}

\def\BibTeX{{\rm B\kern-.05em{\sc i\kern-.025em b}\kern-.08em T\kern-.1667em\lower.7ex\hbox{E}\kern-.125emX}}
\newcommand{\journaladd}[1]{\textcolor{black}{#1}}

\newcommand{\structureadd}[1]{{\color{black}#1}}

\newcommand{\lvmyoadd}[1]{{\color{black}#1}}

\definecolor{MethodRevision}{RGB}{0,0,0}
\definecolor{MethodClarification}{RGB}{0,0,0}
\newcommand{\methodrev}[1]{{\color{MethodRevision}#1}}

\definecolor{TPAMIblue}{HTML}{0014FF}

\hypersetup{
    colorlinks=true,
    linkcolor=TPAMIblue,
    citecolor=TPAMIblue,
    urlcolor=TPAMIblue
}

\begin{document}
\title{Echo4DMesh: Temporally Consistent 4D Cardiac Mesh Reconstruction from Sparse Echocardiography}
\author{Yanan~Liu, Xiaohan~Yuan, Xuan~Yang, Siqi~Xie, Hao~Zhang, Kangjian~He, Hao~Li, Yao~Hao~Teo, Tony~Li, Mark~Y~Chan, Ching~Hui~Sia, Jiayu~Feng, Roger~Foo, Dan~Xu, and Lei~Li\textsuperscript{\Letter}%
\thanks{Y. Liu, X. Yuan, X. Yang, H. Li and L. Li are with the Department of Biomedical Engineering, National University of Singapore, Singapore. Y. Liu is also with the School of Information Science and Engineering, Yunnan University, Kunming, China. X. Yuan is also with Southeast University, Nanjing, China. Y. Liu completed the primary work of this study during the YNU-NUS CSC joint Ph.D. program.}%
\thanks{H. Zhang, K. He, and D. Xu are with the School of Information Science and Engineering, Yunnan University, Kunming, China.}%
\thanks{S. Xie is with Sichuan Cancer Hospital, Chengdu, China.}%
\thanks{M. Y. Chan and C. H. Sia are with Department of Medicine, National University of Singapore, Singapore and Department of Cardiology, National University Heart Centre Singapore, Singapore.}
\thanks{Y. H. Teo and T. Li are with Department of Cardiology, National University Heart Centre Singapore, Singapore.}
\thanks{J. Feng and R. Foo are with Cardiovascular Metabolic Translational Research Programme, Yong Loo Lin School of Medicine, National University of Singapore, Singapore.}
\thanks{\textsuperscript{\Letter}Corresponding author: L. Li (e-mail: \href{mailto:lei.li@nus.edu.sg}{lei.li@nus.edu.sg}).}}

\markboth{Submitted}%
{Liu \MakeLowercase{\textit{et al.}}: Echo4DMesh}

\maketitle

\begin{abstract}
Reconstructing 4D (3D+t) cardiac meshes from sparse echocardiographic (echo) videos enables quantitative analysis of cardiac shape and motion. However, limited observations and uncertain imaging planes make geometry recovery ambiguous, while complex deformation further complicates temporal mesh correspondence. 
We propose Echo4DMesh, an implicit echo based 4D cardiac reconstruction framework that couples patient-specific geometry estimation with temporally consistent mesh generation. Specifically, we use a conditional signed distance function (SDF), pretrained with statistical shape priors, to constrain unobserved anatomy. Then, we propose a SDF-tailored differentiable rendering strategy, jointly refining shape and transducer parameters from 2D masks without paired clinical 3D supervision. To maintain temporal correspondence, we utilize a neural velocity field to propagate a fixed-topology reference mesh, with a radial SDF alignment (RSA) correcting propagated vertices toward the evolving SDF surfaces. Its temporally consistent extension (TC-RSA) integrates bidirectional trajectory fusion and adjacent-frame vertex blending with SDF correction to reduce accumulated drift while preserving surface agreement. Experiments on five synthetic and clinical datasets spanning healthy and pathological anatomy demonstrate reconstruction accuracy, adaptation across disease cohorts, and applicability to whole-heart modeling. Echo4DMesh achieved 98.35\% observed-view Dice on a public clinical benchmark. Patient-matched 3D CT evaluation yielded 90.07\% 3D left-ventricular Dice and 1.81 mm average symmetric surface distance after similarity alignment. Dynamic validation using cine MRI-derived pseudo-echo inputs achieved 94.43\% mean 4D left-ventricular Dice and an absolute ejection-fraction error of 1.62\%. Code will be available upon acceptance of the manuscript for publication via \url{https://github.com/lya19971103/Echo4DMesh}.
\end{abstract}

\begin{IEEEkeywords}
4D cardiac reconstruction, echocardiography, implicit neural representation, differentiable rendering.
\end{IEEEkeywords}

\section{Introduction}

\IEEEPARstart{R}{econstructing} patient-specific 4D (3D+t) cardiac geometry enables quantitative analysis of cardiac morphology~\cite{Chen1994LV,Zhuang2019MvMM,Nascimento2020OneShot}, motion~\cite{yxh,Carneiro2013Tracking,Ye2023SequenceMorph}, and function~\cite{Ouyang2020EchoNet}, providing an anatomical basis for cardiac digital twins~\cite{cai,UltraTwin,CorralAcero2020DigitalTwin} and personalized intervention planning~\cite{Prakosa2018VirtualHeart}. 
Although cardiac computed tomography (CT) and magnetic resonance imaging (MRI) provide detailed anatomical information, their acquisition costs, examination requirements, and, for CT, radiation exposure limit repeated imaging~\cite{journal/MedIA/li2023}. 
Echocardiography (echo) offers an accessible, real-time alternative, but routine 2D acquisitions capture only sparse cross-sections of the heart. 
Recovering complete, temporally coherent cardiac meshes from these observations is therefore an important yet underconstrained reconstruction problem.

The first challenge is \textit{geometric ambiguity under uncertain imaging geometry}. 
Sparse echo views leave substantial portions of the cardiac surface unobserved, allowing multiple 3D shapes to explain the available cross-sections~\cite{UltraTwin}. 
Anatomical priors help constrain these regions, but their ability to represent individual anatomy depends on the variation captured during training, particularly in pathological cases. 
Uncertain transducer placement introduces further ambiguity: discrepancies between a reconstructed cross-section and an observed mask may arise from errors in either the shape or the imaging plane~\cite{MIA25}. 
Patient-specific reconstruction therefore requires adapting anatomical geometry while accounting for uncertainty in the acquisition geometry.
The second challenge is \textit{temporal mesh consistency}. 
A dynamic cardiac reconstruction must capture changing geometry while preserving vertex correspondence and mesh connectivity across frames~\cite{conf/MICCAI/liu2026}. 
Independently extracting meshes from frame-wise implicit surfaces does not establish these correspondences. 
Propagating a reference mesh preserves its vertex indexing and connectivity, but motion estimation errors can accumulate, causing the mesh to drift from the evolving surface. 
Accurate 4D reconstruction thus requires both coherent vertex trajectories and agreement with the geometry reconstructed at each frame.

To address these challenges, we propose Echo4DMesh, an implicit framework for patient-specific 4D cardiac mesh reconstruction based on multi-view echo. 
We address geometric ambiguity using statistical shape model (SSM)-pretrained conditional signed distance functions (SDFs) and an SDF-tailored differentiable rendering strategy that jointly optimizes shape codes and transducer parameters against clinical echo masks, without paired clinical 3D supervision. 
For temporal reconstruction, a neural velocity field propagates a reference mesh with fixed vertex indexing and connectivity, while radial SDF alignment (RSA) corrects deviations from the evolving implicit surfaces. 
We further develop temporally consistent RSA by incorporating bidirectional trajectory fusion and consecutive-frame vertex blending before surface correction, mitigating accumulated drift while maintaining geometric agreement. 
The framework also supports multi-structure reconstruction through structure-specific implicit representations and transducer parameters shared across structures within each view.
Our main contributions are summarized as follows:
\begin{itemize}
    \item We present Echo4DMesh, a novel framework for reconstructing patient-specific 4D cardiac meshes from sparse multi-view echo masks, supporting both single-structure and multi-structure modeling.
    
    \item We formulate patient-specific fitting through SDF-tailored differentiable cross-sectional rendering, jointly refining implicit geometry and transducer parameters without paired clinical 3D supervision.
    
    \item We introduce temporally consistent RSA, which integrates bidirectional trajectory fusion, adjacent-frame vertex blending, and SDF surface correction to improve temporal consistency while preserving reference mesh indexing and connectivity.
    
    \item We evaluate reconstruction accuracy, temporal consistency, and adaptation across disease cohorts using synthetic and clinical datasets, including patient-matched 3D CT references and MRI-derived pseudo-echo sequences.
\end{itemize}

This work extends our preliminary conference study, Echo4DIR~\cite{liu2026Echo4DIR}, which investigated implicit reconstruction of the left ventricular (LV) endocardial surface in healthy subjects. 
The present study introduces new temporally consistent RSA for improved temporal mesh consistency, incorporates disease-specific shape priors for adaptation to pathological anatomy, and extends reconstruction to multiple cardiac structures in a shared coordinate system. 
We further expand the evaluation to healthy and diseased clinical cohorts, patient-matched echo-CT data for end-diastolic reconstruction, and MRI-derived pseudo-echo sequences for 4D reconstruction.

\section{Related Work}

\subsection{Cardiac Reconstruction from Sparse Echocardiography}

Cardiac reconstruction from sparse echo has been explored using different anatomical representations and supervision strategies. 
E-PiVox~\cite{Pix} learned volumetric reconstruction from synthetic echo views, whereas UltraTwin~\cite{UltraTwin} used paired and pseudo-paired US-CT data within a coarse-to-fine generative framework. 
Laumer et al.~\cite{MIA23} learned dynamic cardiac meshes from unpaired echo and mesh sequences; their subsequent work adapted a synthetically pretrained LV model using clinical cross-sectional supervision without paired clinical 3D ground truth~\cite{MIA25}. 
Beyond anatomical modeling, uncertain imaging-plane poses introduced ambiguity between shape and acquisition geometry. 
Chen et al.~\cite{Echo3D2026} addressed this issue by alternating pose optimization and implicit shape fitting, while EchoPOSE~\cite{EchoPOSE2026} combined 6D slice-pose estimation with graph-harmonic LV reconstruction. 
These studies establish the importance of anatomical priors and acquisition geometry in sparse-view reconstruction. 
Echo4DMesh integrated SSM-pretrained implicit shape priors with joint shape and transducer refinement, fitting the observed cross-sections while constraining unobserved anatomy.

\subsection{Implicit Shape Representations and Differentiable Rendering}

Implicit representations encode surfaces as level sets of continuous spatial functions, allowing geometry to be modeled independently of mesh discretization. DeepSDF~\cite{SDF} learned latent-conditioned SDFs for shape representation and surface extraction at flexible resolutions. 
In cardiac modeling, SDF4CHD~\cite{Kong} combined diagnosis-conditioned SDFs with deformation models to capture anatomical variation in congenital heart disease. Fitting such representations to image observations requires a differentiable observation model. 
Soft Rasterizer~\cite{SoftRas2019} enabled image-based optimization of explicit triangle meshes, while differentiable volumetric rendering~\cite{DVR2020} and SDFDiff~\cite{SDFDiff2020} supported optimization of implicit geometry through differentiable rendering.
Echo reconstruction requires matching anatomical cross-sections rather than conventional camera projections. 
Echo4DMesh therefore evaluates conditional SDFs on learnable imaging planes and converts their values into differentiable mask predictions, enabling joint optimization of patient-specific geometry and transducer parameters.

\subsection{Dynamic Reconstruction and Temporal Mesh Consistency}

Dynamic reconstruction must represent evolving geometry while maintaining surface correspondence across time. 
In cardiac imaging, DeepMesh~\cite{DeepMesh2024} estimated vertex-wise LV displacements relative to end diastole from short- and long-axis cine CMR, while Mesh4D~\cite{Mesh4D2025} jointly modeled ventricular shape and motion using continuous deformation fields. 
Implicit approaches similarly coupled geometry and dynamics: occupancy flow~\cite{La1} transported points through a learned velocity field, and Yuan et al.~\cite{yxh} decoupled implicit myocardial shape from temporal deformation for MRI-based reconstruction. 
Independent surface extraction does not establish shared vertex indexing or connectivity across frames, whereas reference-mesh propagation preserves both but can accumulate deviations from the evolving implicit surface. 
Our preliminary work introduced RSA~\cite{liu2026Echo4DIR} to correct propagated vertices toward the current-frame SDF surface. 
Echo4DMesh extended this strategy with temporally consistent RSA, combining bidirectional trajectory fusion and adjacent-frame vertex blending with SDF correction to improve temporal coherence while maintaining surface agreement.

\section{Methodology}

\begin{figure*}[t]
\centering
\includegraphics[width=0.98\textwidth]{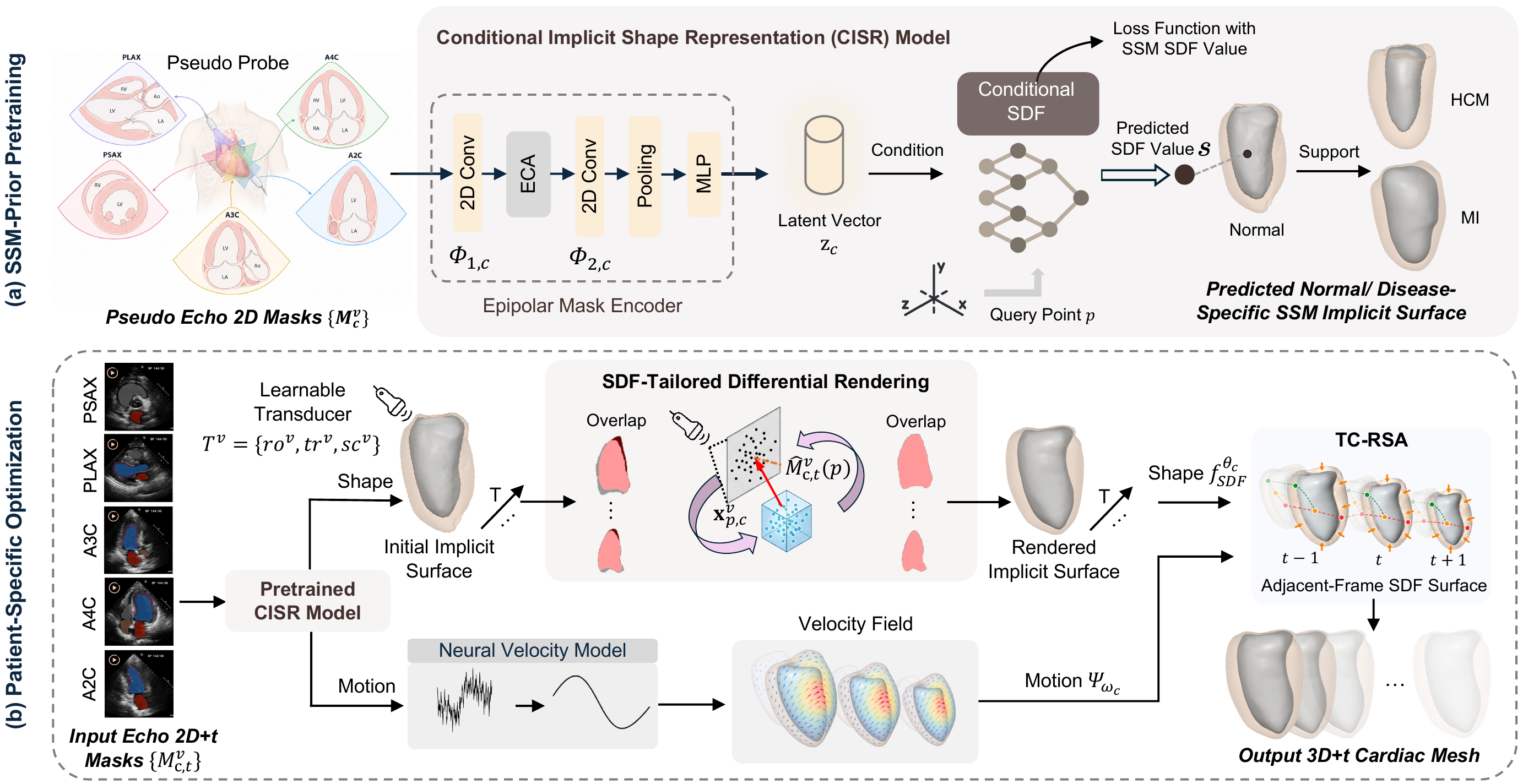}
\caption{Overview of Echo4DMesh.
(a) Anatomical prior learning from statistical shape model (SSM)-derived meshes and synthetic multi-view masks.
(b) Patient-specific reconstruction through joint shape and transducer refinement, followed by velocity-based mesh propagation and temporally consistent signed distance function (SDF) alignment (TC-RSA).
Note that here we only use left ventricular (LV) reconstruction for illustration, but our structure-specific implicit representations also support multi-structure reconstruction in a shared coordinate system. A2C/A3C/A4C: apical two-/three-/four-chamber views; PLAX/PSAX: parasternal long-/short-axis views; HCM: hypertrophic cardiomyopathy; MI: myocardial infarction; ECA: epipolar cross attention.}
\label{fig:method:framework}
\end{figure*}

\subsection{Problem Formulation}

Let $\mathcal C$ denote the set of cardiac structures to be reconstructed, selected from the LV, LV myocardium (MYO), right ventricle (RV), left atrium (LA), and right atrium (RA). 
For structure $c$, let $\mathcal V_c$ contain the available echo views in which it is visible. 
In total, five standard echo views are considered, i,e., apical two-/three-/four-chamber (A2C/A3C/A4C) views and parasternal long-/short-axis (PLAX/PSAX) views.
Given $T$ temporally aligned frames, $M^v_{c,t}$ denotes the binary mask of structure $c$ in view $v\in\mathcal V_c$ at frame $t\in\{0,\ldots,T-1\}$. 
The MYO branch uses the union of the LV cavity and myocardial wall masks, such that its zero-level surface represents the epicardial boundary.
Our objective is to reconstruct patient-specific 4D cardiac geometry without paired clinical 3D supervision:
\begin{equation}
\left\{\mathbf G_{c,t}\right\}
=
\mathcal F_{\mathrm{Echo4DMesh}}
\left(\left\{M^{V_c}_{c,t}\right\}\right),
\label{eq:problem_formulation}
\end{equation}
where $\mathbf G_{c,t}$ is the mesh of structure $c$ at frame $t$ in a shared patient coordinate system. 
The reconstructed meshes should explain the observed cross-sections, provide plausible geometry in unobserved regions, and preserve vertex indexing and structure-specific face connectivity across time.


\subsection{Conditional Implicit Shape Representation}

We first learn a conditional implicit shape model that maps multi-view 2D masks to a 3D cardiac surface, as shown in Fig.~\ref{fig:method:framework} (a).
For each cardiac structure $c$, an epipolar mask encoder (EME) maps the input masks to a latent shape code, and a conditional SDF decoder represents the corresponding 3D geometry.
This model is pretrained on static SSM-derived anatomy and subsequently adapted to temporal echo masks for 4D reconstruction.

Given masks $\{M_c^v\}_{v\in\mathcal V_c^{\mathrm{enc}}}$, where $v$ indexes the views and $\mathcal V_c^{\mathrm{enc}}\subseteq\mathcal V_c$ denotes the encoder input views, a convolutional backbone~\cite{ResNet} with weights shared across views extracts features
$F_c^v=\Phi_{1,c}(M_c^v)\in\mathbb R^{C'\times H'\times W'}$.
Here, $C'$, $H'$, and $W'$ denote the number of channels, height, and width of the feature maps.
For the long-axis branches, epipolar cross attention (ECA)~\cite{ECA} aggregates auxiliary-view information into an anchor view.
The anchor features $F_c^0$ are projected into queries $\mathbf{Q}_c$, while the auxiliary features $F_c^i$ are projected into keys $\mathbf{K}_c$ and values $\mathbf{V}_c$, where $0$ and $i$ index the anchor and auxiliary views, respectively.
For each spatial position $p$ in the anchor feature map, attention is restricted to a vertical window $\mathcal W(p)$ in the auxiliary feature map to accommodate deviations from nominal cross-view alignment:
\begin{equation}
\begin{aligned}
\widehat F_c(p)
&=
F_c^0(p)
+
\sum_{j\in\mathcal W(p)}
a_{p,j}\mathbf{V}_c(j),\\
a_{p,j}
&=
\operatorname{Softmax}_{j\in\mathcal W(p)}
\left(
\frac{\mathbf{Q}_c(p)^\top \mathbf{K}_c(j)}{\tau}
\right),\\
\mathbf z_c
&=
\operatorname{MLP}_c
\left(
\operatorname{AvgPool}
\left(\Phi_{2,c}(\widehat F_c)\right)
\right),
\end{aligned}
\label{eq:latent_z}
\end{equation}
where $j$ indexes positions within $\mathcal W(p)$, $a_{p,j}$ is the attention weight assigned to auxiliary position $j$ for anchor position $p$, and $\tau$ is the attention temperature.
The softmax is normalized over the window, and the anchor features are updated successively using the auxiliary long-axis views.
The fused feature map $\widehat F_c$ is then transformed by the convolutional module $\Phi_{2,c}$, spatial average pooling, and an multilayer perceptron (MLP) to produce the shape code $\mathbf z_c\in\mathbb R^L$, where $L$ is the latent-space dimension.
The complete encoder is denoted by
$\mathbf z_c=\mathrm{EME}_c(\{M_c^v\}_{v\in\mathcal V_c^{\mathrm{enc}}})$.

Conditioned on $\mathbf z_c$, an MLP decoder represents the 3D geometry using an SDF~\cite{SDF}:
\begin{equation}
s_c(\mathbf x)
=
f_{\mathrm{SDF}}^{\theta_c}
(\mathbf z_c,\mathbf x),
\qquad
\mathbf x\in\Omega_c,
\label{eq1}
\end{equation}
where $\theta_c$ denotes the decoder parameters, $\mathbf x\in\mathbb R^3$ is a spatial query point, and $\Omega_c\subset\mathbb R^3$ is the query domain.
The scalar output $s_c(\mathbf x)$ approximates the signed Euclidean distance from $\mathbf x$ to the nearest surface boundary, with negative values inside and positive values outside.
The surface is defined by the zero-level set
$\mathcal S_c=\{\mathbf x\in\Omega_c\mid s_c(\mathbf x)=0\}$,
providing a continuous geometric representation independent of a fixed mesh discretization.

The EME and SDF decoder are jointly pretrained using meshes sampled from normal and disease-specific SSMs and their corresponding pseudo multi-view 2D masks.
Each pretraining sample represents a static 3D anatomy rather than a temporal sequence.
For subsequent patient-specific 4D reconstruction, the pretrained encoder initializes a shape code at each frame $t$ from the observed masks:
$\mathbf z_{c,t}=\mathrm{EME}_c(\{M^v_{c,t}\}_{v\in\mathcal V_c^{\mathrm{enc}}})$.
The shared decoder then represents the geometry at each frame as
$s_{c,t}(\mathbf x)=f_{\mathrm{SDF}}^{\theta_c}(\mathbf z_{c,t},\mathbf x)$.
These frame-specific representations provide the initialization and anatomical prior for the subsequent patient-specific adaptation and temporal modeling.

\subsection{Self-Supervised SDF-Tailored Differentiable Rendering}

For patient-specific reconstruction, the pretrained shape model must adapt to clinical 2D echo masks that may differ from the anatomy represented by the SSM priors.
We therefore introduce an SDF-tailored differentiable rendering (DR) strategy to fit patient-specific geometry to the observed masks while accounting for uncertain probe placement, as shown in Fig.~\ref{fig:method:framework} (b). 
Specifically, we parameterize the transducer geometry of each view $v$ via a learnable spatial transformation $T^v=\{ro^v,tr^v, sc^v\}$, denoting rotation, translation, and scaling, respectively.
A sampled 2D pixel $p=(x,y)$ on the image plane is embedded as $\bar p=[x,y,0]^\top$ and mapped to the shared patient coordinate system. 
For multi-structure reconstruction, $K_c$ maps decoder coordinates to a common reference system, and $H_c$ adjusts the placement of structure $c$ for the patient:
\begin{equation}
\begin{aligned}
\mathbf y^v_{p}
&=
T^v(\bar p)
=
ro^v\operatorname{diag}(sc^v)\bar p+ tr^v,\\
x^v_{p,c}
&=
(H_c\circ K_c)^{-1}(\mathbf y^v_{p}).
\end{aligned}
\label{eq:structure_query}
\end{equation}
The imaging transformation $T^v$ is shared across frames and structures within each view, while $H_c$ is shared across frames. 
For isolated-structure reconstruction, $K_c$ and $H_c$ may be identities.

To compare the implicit geometry with the observed 2D masks, we convert the SDF value at each query point $\mathbf x^v_{p,c}$ into a foreground probability. A hard inside-outside test would produce binary predictions with zero gradients almost everywhere, preventing gradient-based refinement of shape and transducer parameters. We instead use a scaled sigmoid to render a soft cross-sectional mask:
\begin{equation}
\widehat M^v_{c,t}(p)
=
\frac{1}{
1+\exp\left(
\alpha s_{c,t}(\mathbf x^v_{p,c})
\right)},
\label{eq:rendering}
\end{equation}
where $\alpha>0$ controls the sharpness of the foreground-background transition.
This differentiable mapping allows discrepancies between rendered and observed masks to jointly update the shape codes and transducer parameters for patient-specific adaptation.

\subsection{Temporally Consistent Mesh Reconstruction via Neural Velocity Modeling}

The implicit representation describes cardiac geometry at each frame but does not establish vertex correspondence across time. 
We therefore model motion using a continuous neural field that transports a reference mesh through the sequence:
\begin{equation}
\mathbf v_{c,t}(\mathbf x)
=
\Psi_{\omega_c}
\left(\gamma(\mathbf x),\gamma(u_t)\right),
\qquad
u_t=t/T,
\label{eq:structure_velocity}
\end{equation}
where $\Psi_{\omega_c}$ is a MLP and $\gamma$ denotes sinusoidal positional encoding~\cite{NerF}. 
The inter-frame time step is absorbed into the network output, allowing $\mathbf v_{c,t}$ to be applied directly as an incremental displacement. 
The field is learned through bidirectional SDF transport between adjacent frames, as defined in Eq.~\eqref{eq:structure_transport}. 
During temporal reconstruction, the refined shape codes $\widetilde{\mathbf z}_{c,t}$ define the evolving surfaces through
$s_{c,t}(\mathbf x)=f_{\mathrm{SDF}}^{\theta_c}
(\widetilde{\mathbf z}_{c,t},\mathbf x)$.

We extract a reference mesh $(\mathcal X_{c,0},\mathcal F_c)$ from the initial implicit surface using marching cubes (MC)~\cite{Marching}. 
Subsequent meshes are generated by propagating its vertices while retaining their indexing and face connectivity. 
Specifically, each vertex is updated as
$\widehat{\mathbf x}_{c,t}
=
\mathbf x_{c,t-1}
+
\mathbf v_{c,t-1}(\mathbf x_{c,t-1})$.
Motion estimation errors may cause these propagated vertices to deviate from the current-frame surface. 
RSA compensates for this deviation through a signed-distance correction along the local surface normal:
\begin{equation}
\mathbf x_{c,t}
=
\widehat{\mathbf x}_{c,t}
-
s_{c,t}(\widehat{\mathbf x}_{c,t})
\frac{
\nabla_{\mathbf x}s_{c,t}(\widehat{\mathbf x}_{c,t})
}{
\left\|
\nabla_{\mathbf x}s_{c,t}(\widehat{\mathbf x}_{c,t})
\right\|_2
}.
\label{eq:sdf_alignment}
\end{equation}
Here, the SDF value determines the signed correction magnitude, and its normalized gradient specifies the direction. 
This update moves propagated vertices toward the reconstructed zero-level surface without altering mesh connectivity.

Although RSA improves frame-wise surface agreement, sequential propagation can still accumulate errors and produce local trajectory fluctuations. 
We introduce a new temporally consistent-RSA (TC-RSA) to incorporate temporal context before surface correction. 
Let $\mathcal R_{c,t}$ denote the vertex-wise RSA operator in Eq.~\eqref{eq:sdf_alignment}, and let
$\mathcal X^\rightarrow_{c,t}$ and
$\mathcal X^\leftarrow_{c,t}$
denote the forward and backward RSA trajectories with shared reference vertex indexing. 
We first fuse corresponding vertex positions:
\begin{equation}
\begin{aligned}
\mathcal P_{c,t}
&=
(1-\beta_t)\mathcal X^\rightarrow_{c,t}
+
\beta_t\mathcal X^\leftarrow_{c,t},\\
\beta_t&=t/(T-1).
\end{aligned}
\label{eq:trirsa_bidirectional}
\end{equation}
The time-dependent weight shifts the contribution from the forward trajectory toward the backward trajectory over the sequence, reducing reliance on a single propagation direction.
For each interior frame, corresponding vertex positions are then blended across three consecutive frames:
\begin{equation}
\begin{aligned}
\overline{\mathcal P}_{c,t}
&=
\lambda\mathcal P_{c,t-1}
+
(1-2\lambda)\mathcal P_{c,t}
+
\lambda\mathcal P_{c,t+1},\\
\mathcal X^{\mathrm{TC}}_{c,t}
&=
\mathcal R_{c,t}
(\overline{\mathcal P}_{c,t}),
\qquad 1\leq t\leq T-2,
\end{aligned}
\label{eq:trirsa}
\end{equation}
where $\lambda$ weights the neighboring frames. Temporal blending suppresses local fluctuations but may displace vertices from the current surface. We therefore recompute SDF values and gradients at the blended positions and apply RSA to recover surface agreement. The first frame retains the reference vertices,
$\mathcal X^{\mathrm{TC}}_{c,0}=\mathcal X_{c,0}$,
while the final frame receives RSA correction without three-frame blending:
$\mathcal X^{\mathrm{TC}}_{c,T-1}
=
\mathcal R_{c,T-1}(\mathcal P_{c,T-1})$.
Note that TC-RSA requires no additional network parameters or training losses and preserves the reference mesh indexing and connectivity. The resulting meshes are expressed in the shared patient coordinate system as
\begin{equation}
\mathbf G_{c,t}
=
\left(
(H_c\circ K_c)(\mathcal X^{\mathrm{TC}}_{c,t}),
\mathcal F_c
\right).
\label{eq:output_mesh}
\end{equation}

\subsection{Model Optimization and Loss Functions}

Echo4DMesh combines supervised anatomical prior learning with patient-specific fitting to clinical masks. The former learns a shape representation from SSM-derived anatomy, while the latter adapts geometry and imaging parameters to individual observations and estimates motion through implicit surface transport. We omit the structure index $c$ below; for multi-structure reconstruction, losses are summed over structures using their available views.

The encoder and SDF decoder are pretrained using SSM-derived meshes and simulated cross-sectional masks~\cite{MIA23,MIA25}, including normal and disease-specific anatomical distributions. Given sampled query points $\mathbf x\in\Omega$, we minimize
\begin{equation}
\begin{aligned}
\mathcal L_{\mathrm{pt}}
={}&
\frac{1}{|\Omega|}
\sum_{\mathbf x\in\Omega}
\left|
f_{\mathrm{SDF}}^\theta(\mathbf z,\mathbf x)
-
s_{\mathrm{gt}}(\mathbf x)
\right| +
\lambda_z\|\mathbf z\|_2^2,
\end{aligned}
\label{eq:pretrain_loss}
\end{equation}
where $s_{\mathrm{gt}}$ is the ground-truth SDF from the SSM mesh. The reconstruction term supervises the implicit geometry, and
$\lambda_z$ is a balancing parameter.
To represent pathological anatomy, we also learn shape priors from disease-specific SSMs for HCM and MI using the same pretraining objective in Eq.~\eqref{eq:pretrain_loss}. These priors capture disease-related morphological variation, while subsequent fitting to patient-specific clinical masks refines the geometry to match individual anatomy. 
Adaptation therefore follows the same reconstruction procedure across cohorts, without requiring paired clinical 3D supervision.

For patient-specific reconstruction, the encoder initializes shape codes from clinical masks. The codes and transducer parameters are subsequently refined by matching rendered cross-sections to the observations. We combine pixel-wise binary cross-entropy (BCE) loss $L_{\mathrm{BCE}}$ with Dice loss $L_{\mathrm{Dice}}$ to account for both local classification and regional overlap:
\begin{equation}
\begin{aligned}
\mathcal L_{\mathrm{render}}
=
\frac{1}{N_{view}}\sum_{v=1}^{N_{view}}
\Bigg[
&\frac{1}{|\mathcal P|}
\sum_{p\in\mathcal P}
\mathcal L_{\mathrm{BCE}}
\left(\widehat M^v(p),M^v(p)\right)\\
&+
 \lambda_{\mathrm{Dice}} \mathcal L_{\mathrm{Dice}}
\left(\widehat M^v,M^v\right)
\Bigg],
\end{aligned}
\label{eq:hybrid_loss}
\end{equation}
where $N_{view}$ is the number of fitting views, $\mathcal P$ is the sampled pixel set, and $\lambda_{\mathrm{Dice}}$ is
a balancing parameter. The frame index is omitted for clarity.

The motion field is learned from the evolving implicit surfaces. Following particle advection~\cite{La1} and level-set evolution~\cite{La2}, we encourage points near a source surface to remain near the adjacent target surface after transport. Applying this constraint in both temporal directions gives
\begin{equation}
\begin{aligned}
\mathcal L_{\mathrm{transport}}
={}&
\mathbb E_{\mathbf x_t}
\left[
w_t(\mathbf x_t)
\left|
s_{t+1}
\left(
\mathbf x_t+\mathbf v_t(\mathbf x_t)
\right)
\right|^2
\right]\\
&+
\mathbb E_{\mathbf x_{t+1}}
\left[
w_{t+1}(\mathbf x_{t+1})
\left|
s_t
\left(
\mathbf x_{t+1}
-
\mathbf v_{t+1}(\mathbf x_{t+1})
\right)
\right|^2
\right]\\
&+
\mathcal \lambda_{smooth}L_{\mathrm{smooth}},
\end{aligned}
\label{eq:structure_transport}
\end{equation}
where $w_t(\mathbf x)=\exp(-|s_t(\mathbf x)|)$ emphasizes samples near the source zero-level surface. The first two terms penalize residual signed distances after forward and backward transport, respectively. $\lambda_{smooth}$ is a balancing parameter. To discourage unnecessarily large displacements, we use
\begin{equation}
\mathcal L_{\mathrm{smooth}}
=
\mathbb E_{\mathbf x\sim\mathcal U(\Omega)}
\left[
\|\mathbf v_t(\mathbf x)\|_2^2
\right],
\label{eq:motion_regularization}
\end{equation}
Together, these terms provide surface-based motion supervision without ground-truth vertex trajectories.

\section{Experiments and Results}

We evaluate Echo4DMesh on five synthetic and clinical datasets encompassing healthy subjects and patients with MI or HCM. 
Experiments assess reconstruction accuracy, adaptation across disease cohorts, and the contributions of individual components under matched input-view configurations and evaluation protocols. 
Patient-matched CT and MRI-derived pseudo-echo inputs provide complementary evaluations of static geometry and dynamic reconstruction, respectively. 
We further assess extension from LV to whole-heart reconstruction.

\subsection{Datasets and Preprocessing}

\textbf{1) SSM.}
Following~\cite{MIA23,MIA25}, we generated 10,000 synthetic LV meshes from an 18-mode SSM, using 9,000 for training and 1,000 for testing. Multi-view 2D masks were obtained by intersecting each mesh with simulated echo imaging planes. 
For each mesh, 100,000 spatial points were sampled to compute reference signed distances. 
These data support conditional SDF pretraining and geometric evaluation against complete 3D references.
Disease-specific SSMs were constructed from 30 HCM and 30 MI subjects in ACDC~\cite{bernard2018acdc}. 
For each subject, we identified its end diastole (ED) and end systolic (ES) phases, and manually annotated the ED-ES-ED cycle frame by frame. 
Geometry was recovered in physical coordinates using voxel spacing and represented using the reference topology and coordinate conventions of the normal SSM. 
Principal component analysis was performed separately for each disease to construct shape spaces, from which synthetic meshes were sampled. Simulated multi-view masks and sampled signed distances provided paired inputs and supervision for conditional SDF pretraining. 

\textbf{2) Control Cohort (CC).}
CC contained transthoracic echo data from 144 healthy subjects~\cite{MIA25}. 
Sequences underwent quality filtering, alignment to end diastole, and resizing to $224\times224$. 
LV blood-pool masks for the A2C, A3C, and A4C views were obtained using nnU-Net~\cite{nnunet}. 
In the absence of paired 3D references, evaluation measured agreement between reconstructed cross-sections and the echo masks.

\textbf{3) PICMAN.}
PICMAN included paired 2D echo and 3D CT images from 91 normal subjects without the target cardiac diseases from National University Hospital (NUH), Singapore. 
A2C, A3C, and A4C mask sequences were used for three-view, held-out-view, and dynamic reconstruction experiments. 
For CT-based evaluation, we randomly selected 60 subjects with complete end-diastolic masks in these views and patient-matched CT-derived LV segmentations. 
CT references were used exclusively for registration and evaluation, without contributing to LV reconstruction or model selection. 
Subjects with the required additional view annotations were also included in whole-heart reconstruction experiments in Sec.~\ref{sec:exp:whole_heart}.

\textbf{4) NUH-MI.}
This cohort included 36 echo cases and 103 MRI cases from post-MI patients at NUH, without patient-level pairing between modalities.
A2C, A3C, and A4C mask sequences were used to evaluate three-view reconstruction, held-out-view prediction, and dynamic reconstruction under pathological changes in cardiac geometry and motion.

\textbf{5) NUH-HCM.}
This cohort comprised 62 HCM patients from NUH and was used to assess reconstruction of hypertrophic anatomy and adaptation to disease-related morphological variation. 
Cases with complex geometry and motion further supported comparisons of RSA and TC-RSA.

\begin{figure}[t]
    \centering
    \includegraphics[width=\columnwidth]{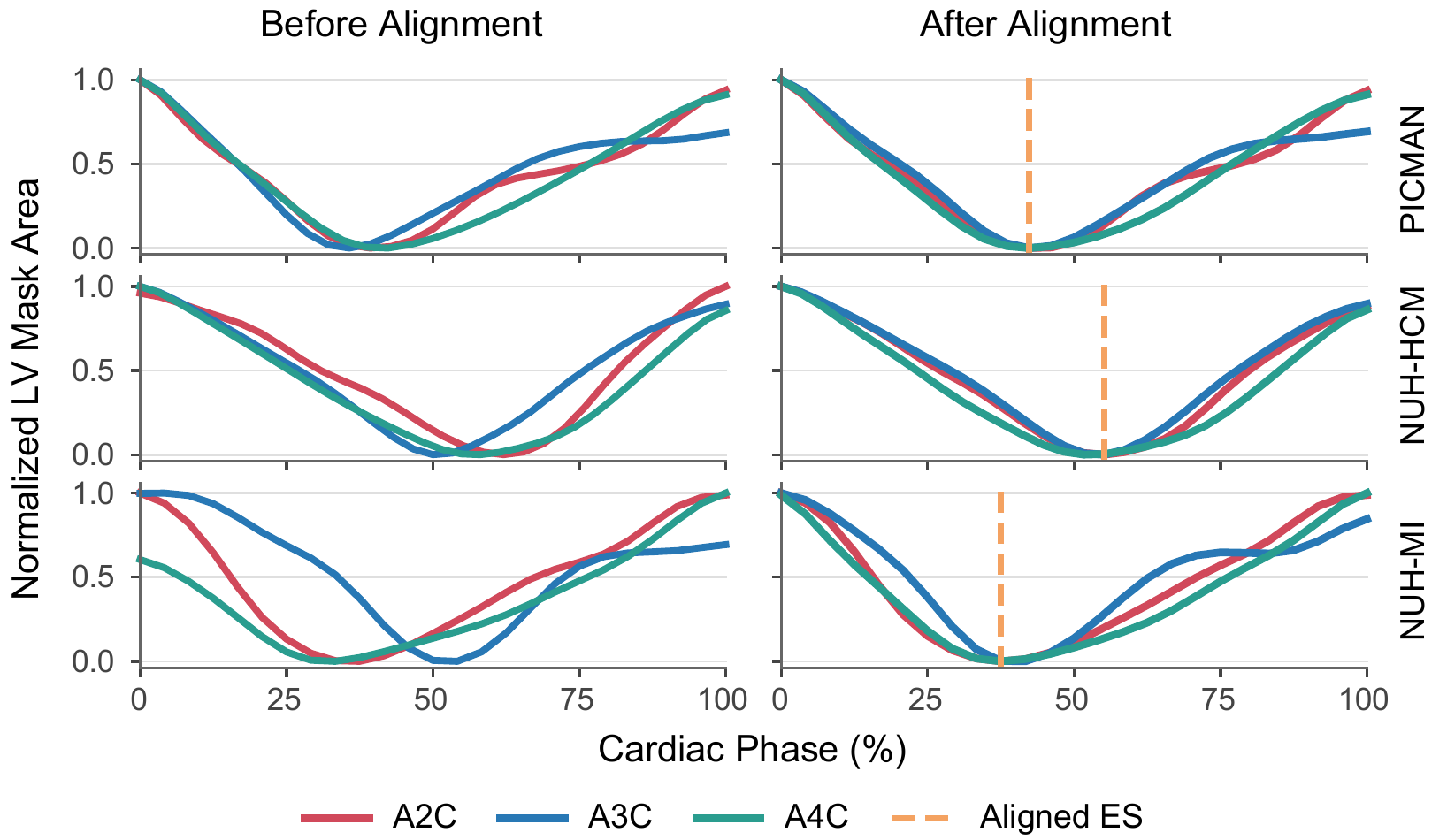}
    \caption{Cross-view temporal alignment in three representative cases. Normalized LV mask-area curves for A2C, A3C, and A4C are shown before and after synchronization to a common end-systolic (ES) phase.}
    \label{fig:temporal_alignment}
\end{figure}

For PICMAN, NUH-MI, and NUH-HCM, we synchronized the views using their LV mask-area curves. 
Each sequence is shifted to begin at ED, and a common ES phase is determined from the area changes across views. 
Systolic and diastolic intervals are then resampled separately to produce temporally aligned ED-ES-ED sequences. 
Representative examples are shown in Fig.~\ref{fig:temporal_alignment}.

\subsection{Evaluation Metrics}

On the synthetic SSM benchmark, geometric accuracy was assessed using mean absolute error (MAE), Hausdorff distance (HD), Chamfer distance (CD), reported in millimeters. 
On clinical echo datasets, Dice and intersection over union (IoU) measured agreement between reconstructed cross-sections and segmentation masks. 
Dynamic scores were computed per frame and averaged within each subject.
Held-out-view evaluation assessed cross-sectional agreement in views excluded from shape fitting. 
Held-out masks were used for preprocessing and transducer pose calibration, but were excluded from shape encoding and rendering losses. The same calibration protocol was applied to all methods. 
ECA was bypassed for single-view inputs, while multi-view feature fusion used only the fitting views. 
This protocol evaluated geometric completion under calibrated imaging geometry.

For patient-matched echo-CT evaluation, reconstructed meshes were aligned to CT references using global similarity registration, allowing translation, rotation, and uniform scaling without local deformation.
Shape agreement was assessed using 3D Dice, average symmetric surface distance (ASSD), and 95th-percentile Hausdorff distance (HD95).
These metrics evaluated geometry after alignment.
For MRI-based dynamic evaluation, reconstructions from MRI-derived pseudo-echo masks were compared with the corresponding MRI mesh sequences.
We measured frame-averaged 3D Dice, absolute ejection-fraction (EF) error, volume gap, and velocity RMSE.
Velocity RMSE quantified errors in the temporal derivative of displacement relative to ED and was normalized by the MRI-derived ED LV long-axis length, $L_0$, with units of $\%L_0$/cycle.
This evaluation used MRI-derived simulated cross-sections to assess dynamic reconstruction, given the limited availability of paired clinical echo-MRI acquisitions.
The volume gap measured the discrepancy between the normalized volume curves of a reconstructed mesh sequence and a reference mesh sequence:
\begin{equation}
V_{gap}=\frac{1}{T}\sum_{t=0}^{T-1}
\left|
\frac{V_t^{pre}}{V_0^{pre}}
-
\frac{V_t^{\mathrm{ref}}}
     {V_0^{\mathrm{ref}}}
\right|,
\label{eq:volume_gap}
\end{equation}
where $V_t^{pre}$ denoted the volume of the predicted mesh, and $V_t^{\mathrm{ref}}$ denoted the reference mesh volume at frame $t$.
For mesh-to-SDF evaluation, the reference was obtained by independently applying MC~\cite{Marching} to the optimized SDF; for MRI-based evaluation, it was the corresponding MRI mesh.

\subsection{\structureadd{Implementation Details}}
Our framework was implemented in PyTorch on an RTX 4090 GPU. The encoders $\Phi_1$ and $\Phi_2$ use ResNet-18 with feature dimensions $(64,128)$ and $(256,512)$, respectively. For $224\times224$ input masks, ECA operated on $28\times28$ feature maps with four attention heads, a $5\times1$ window (height $\times$ width), and zero padding at vertical boundaries. The SDF network followed~\cite{SDF}, and $\Psi_\omega$ was a four-layer MLP with positional encoding.

During patient-specific 4D reconstruction stage, we optimized the first frame as the anchor for 1,500 steps, alternating shape and probe optimization at an 8:2 time ratio. We then froze the probe parameters and optimize shape and motion for 2,000 steps with random frame selection. Adam learning rates were $5\times10^{-3}$, $5\times10^{-4}$, and $1\times10^{-3}$ for the probe, latent shape, and velocity field, respectively. We set $\lambda_{z}=10^{-6}$, $\lambda_{Dice}=0.5$, and $\lambda_{smooth}=0.001$. Rendering sampled 4,096 points per iteration, with $\alpha$ increasing from 20 to 150. \methodrev{TC-RSA used a neighboring-frame weight of $\lambda=0.125$ in Eq.~\eqref{eq:trirsa}.} 3D and 4D reconstruction took approximately 30~s and 2~min, respectively. Further experimental details and parameter settings will be presented in our repository.

For the whole-heart reconstruction, each structure had a separate pretrained encoder and SDF decoder, with transducer parameters shared across structures within each view. LV, MYO, and LA used A4C/A2C/A3C masks; RV used A4C/PLAX/PSAX; and RA used A4C/PSAX.  Fixed transforms $K_c$ were obtained by similarity-registering pretrained first-frame surface predictions from PICMAN dataset to the corresponding structures of a common whole-heart SSM sample. Structure poses $H_c$ were initialized to identity and calibrated from echo masks through bounded rotations and translations. Calibration used 15 alternating cycles of 50 transducer-parameter updates and 50 structure-pose updates. We then fixed both parameter sets and perform 1,200 joint shape-motion optimization steps, sharing randomly sampled phases across structures for rendering supervision. Learning rates were set as $5\times10^{-5}$ for shape codes and $5\times10^{-4}$ for velocity networks.

\begin{table}[!t]
\centering
\caption{Summary of quantitative results of 3D geometric reconstruction accuracy on the synthetic SSM dataset.}
\label{tab:reconstruction_results}

\fontsize{7pt}{7pt}\selectfont
\setlength{\tabcolsep}{2pt}
\renewcommand{\arraystretch}{1.12}
\begin{tabular}{l ccc}
\toprule
\multicolumn{1}{c}{\multirow[c]{2}{*}[-0.75ex]{\textbf{Method}}}
& \multicolumn{3}{c}{\textbf{SSM Dataset}} \\
\cmidrule(lr){2-4}
& MAE (mm) $\downarrow$
& HD (mm) $\downarrow$
& CD (mm) $\downarrow$ \\
\midrule
E-PiVox~\cite{Pix}
& 3.41 $\pm$ 0.85
& 13.04 $\pm$ 4.52
& 7.34 $\pm$ 1.88 \\
Laumer et al.~\cite{MIA25}
& 2.61 $\pm$ 0.78
& 7.98 $\pm$ 1.43
& 5.61 $\pm$ 0.89 \\
\textbf{Ours}
& \textbf{1.15 $\pm$ 0.13}
& \textbf{5.49 $\pm$ 0.54}
& \textbf{2.84 $\pm$ 0.85} \\
\bottomrule
\end{tabular}
\end{table}

\begin{table*}[t]
\centering
\caption{Summary of the quantitative evaluation results of 2D mask overlap on four clinical datasets. Results are patient-level mean ± SD after averaging each metric over three views.}
\label{tab:clinical_reconstruction_results}
\resizebox{0.94\textwidth}{!}{
\begin{tabular}{l cc cc cc cc}
\toprule
\multicolumn{1}{c}{\multirow[c]{2}{*}[-0.75ex]{\textbf{Method}}}
& \multicolumn{2}{c}{\textbf{CC}}
& \multicolumn{2}{c}{\textbf{PICMAN}}
& \multicolumn{2}{c}{\textbf{NUH-HCM}}
& \multicolumn{2}{c}{\textbf{NUH-MI}} \\
\cmidrule(lr){2-3} \cmidrule(lr){4-5}
\cmidrule(lr){6-7} \cmidrule(lr){8-9}
& Dice (\%) $\uparrow$ & IoU (\%) $\uparrow$
& Dice (\%) $\uparrow$ & IoU (\%) $\uparrow$
& Dice (\%) $\uparrow$ & IoU (\%) $\uparrow$
& Dice (\%) $\uparrow$ & IoU (\%) $\uparrow$ \\
\midrule

E-PiVox~\cite{Pix}
& 87.35 $\pm$ 3.48 & 77.97 $\pm$ 4.02
& 81.66 $\pm$ 5.67 & 69.54 $\pm$ 7.40
& 60.33 $\pm$ 10.93 & 44.67 $\pm$ 11.12
& 74.45 $\pm$ 8.27 & 60.18 $\pm$ 9.81 \\

Laumer et al.~\cite{MIA25}
& 90.51 $\pm$ 2.72 & 84.76 $\pm$ 3.86
& 92.31 $\pm$ 2.06 & 85.78 $\pm$ 3.55
& 89.23 $\pm$ 3.58 & 80.74 $\pm$ 5.81
& 92.24 $\pm$ 2.36 & 85.61 $\pm$ 4.02 \\

\textbf{Ours}
& \textbf{98.35 $\pm$ 0.42}
& \textbf{96.75 $\pm$ 0.80}
& \textbf{98.12 $\pm$ 0.69}
& \textbf{96.31 $\pm$ 1.32}
& \textbf{94.68 $\pm$ 2.81}
& \textbf{90.03 $\pm$ 4.86}
& \textbf{97.89 $\pm$ 0.92}
& \textbf{95.88 $\pm$ 1.74} \\

\bottomrule
\end{tabular}}
\end{table*}

\begin{figure}[t]
\centering
\includegraphics[width=0.5\textwidth]{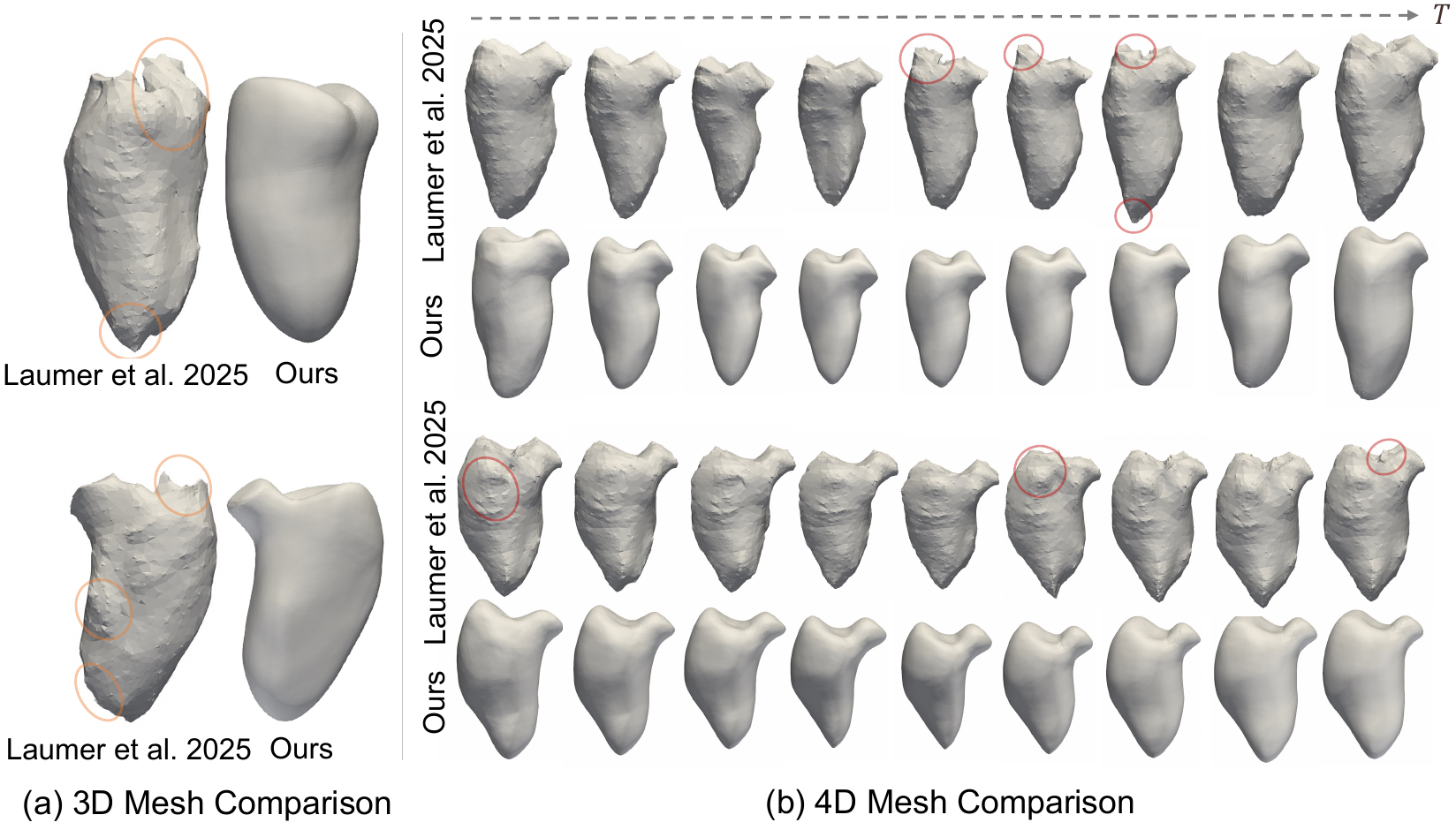}
\caption{3D/4D visualization of reconstructed LV geometry on the public CC dataset via the proposed method and Laumer et al.~\cite{MIA25}.}
\label{fig:result:3d_comparison_CC}
\end{figure}

\begin{figure}[t]
    \centering
    \includegraphics[width=0.5\textwidth]{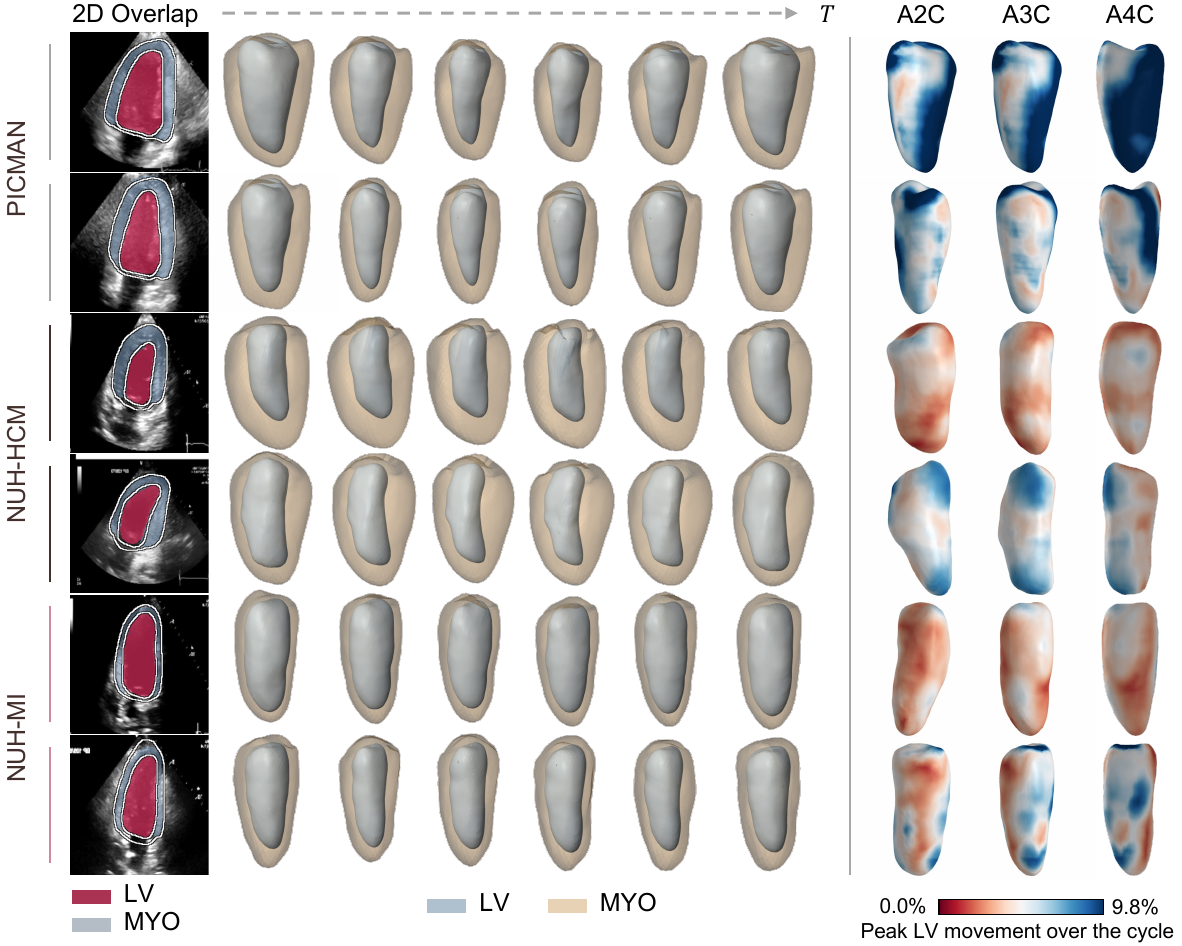}
    \caption{\lvmyoadd{4D LV-MYO reconstruction and LV surface displacement across different cohorts. The first column overlays reconstructed mesh cross-sections on frame-0 reference masks, followed by six cardiac phases. The rightmost columns show the frame-0 LV surface colored by peak displacement over the cycle after rigid-motion correction, viewed from A2C, A3C, and A4C orientations. Displacement was normalized by the frame-0 LV long-axis length.}}
    \label{fig:result:cross_cohort_4d_mesh}
\end{figure}

\subsection{Comparison with State-of-the-Art Methods} \label{sec:exp:comparison}

We compared Echo4DMesh with E-PiVox~\cite{Pix} and Laumer et al.~\cite{MIA25}, first evaluating 3D geometric accuracy on the synthetic SSM dataset and then cross-sectional agreement on four clinical cohorts. 
On the SSM dataset, where complete reference surfaces are available, Echo4DMesh achieved the lowest errors across all three metrics (Table~\ref{tab:reconstruction_results}). 
Compared with Laumer et al.~\cite{MIA25}, MAE decreased from 2.61 to 1.15~mm, HD from 7.98 to 5.49~mm, and CD from 5.61 to 2.84~mm, indicating more accurate recovery of 3D geometry from sparse 2D observations.
On the clinical datasets, Echo4DMesh achieved the highest Dice and IoU across all four cohorts after patient-specific optimization (Table~\ref{tab:clinical_reconstruction_results}). 
Dice scores were 98.35\%, 98.12\%, 94.68\%, and 97.89\% on CC, PICMAN, NUH-HCM, and NUH-MI, respectively. 
Relative to Laumer et al.~\cite{MIA25}, Dice increased by 5.45\%-7.84\% and IoU by 9.29\%-11.99\%. 
These results demonstrate closer agreement between reconstructed mesh cross-sections and clinical segmentation masks.

Figure~\ref{fig:result:3d_comparison_CC} presented qualitative comparisons with Laumer et al.~\cite{MIA25} on the CC dataset. 
The circled regions in Fig.~\ref{fig:result:3d_comparison_CC} (a) highlighted local protrusions and uneven contours in the baseline reconstructions, whereas Echo4DMesh produced smoother LV surfaces. 
This difference potentially reflected the underlying shape representations: Laumer et al. directly predicted discrete mesh vertex positions using a graph-based decoder, whereas Echo4DMesh reconstructed a continuous implicit surface constrained by an SSM-derived anatomical prior. 
This formulation provided a global shape constraint when fitting sparse observations, which potentially limited local surface irregularities. Fig.~\ref{fig:result:3d_comparison_CC} (b) further showed smoother surfaces and more coherent shape changes across the displayed cardiac phases. 
The contributions of surface alignment and temporal refinement were evaluated separately in the ablation study (see Sec.~\ref{sec:exp:ablation_study}).

\subsection{Cross-Disease Adaptation} \label{sec:exp:cross-disease}

Table~\ref{tab:clinical_reconstruction_results} showed improved cross-sectional agreement across all evaluated clinical cohorts. Fig.~\ref{fig:result:cross_cohort_4d_mesh} further illustrated adaptation to anatomical and motion variations in representative healthy, HCM, and MI cases. 
The initial-frame overlays showed close alignment between reconstructed cross-sections and reference masks despite differences in cavity shape and myocardial thickness. Across the six displayed phases, the LV-MYO meshes maintained smooth surfaces while representing contraction and relaxation with different amplitudes. 
The HCM reconstructions captured thicker myocardium than the healthy cases, together with distinct degrees of cavity contraction, indicating that the model accommodated both structural variation and individual motion patterns.

The peak LV displacement maps summarized regional motion over the full cycle, allowing spatial differences to be assessed beyond the selected mesh snapshots. 
Displacements were corrected for rigid motion and normalized by the initial-frame LV long-axis length to facilitate comparison across subjects. 
The healthy cases showed relatively extensive high-displacement regions, whereas the HCM cases exhibited differing displacement magnitudes despite their shared hypertrophic morphology. 
The MI cases showed either broadly low displacement or a mixture of low- and higher-displacement regions, illustrating the capacity of the proposed model to represent heterogeneous motion rather than imposing uniform contraction. 
Although correspondence with scar location remained unverified, these regional patterns complemented the mask overlap and mesh sequences in supporting adaptation to diverse cardiac shapes and motion patterns.

\begin{table*}[t]
\centering
\caption{Patient-specific adaptation and cross-view consistency across four clinical datasets. DR denoted differentiable rendering-driven shape adaptation; w/o DR retained pose calibration but omitted shape adaptation. In the held-out-view setting, A3C+A4C $\rightarrow$ A2C denoted shape adaptation using A3C and A4C, with A2C used only for pose calibration and evaluation.}
\label{tab:dr_cross_view}
\resizebox{0.94\textwidth}{!}{
\begin{tabular}{l cc cc cc cc}
\toprule
\multicolumn{1}{c}{\multirow[c]{2}{*}[-0.75ex]{\textbf{Method}}}
& \multicolumn{2}{c}{\textbf{CC}}
& \multicolumn{2}{c}{\textbf{PICMAN}}
& \multicolumn{2}{c}{\textbf{NUH-HCM}}
& \multicolumn{2}{c}{\textbf{NUH-MI}} \\
\cmidrule(lr){2-3}
\cmidrule(lr){4-5}
\cmidrule(lr){6-7}
\cmidrule(lr){8-9}

& Dice (\%) $\uparrow$ & IoU (\%) $\uparrow$
& Dice (\%) $\uparrow$ & IoU (\%) $\uparrow$
& Dice (\%) $\uparrow$ & IoU (\%) $\uparrow$
& Dice (\%) $\uparrow$ & IoU (\%) $\uparrow$ \\
\midrule

Ours $w/o$ DR
& 88.10 $\pm$ 3.08 & 78.86 $\pm$ 4.42
& 87.40 $\pm$ 2.45 & 77.72 $\pm$ 3.68
& 76.73 $\pm$ 4.87 & 63.28 $\pm$ 6.09
& 86.06 $\pm$ 3.31 & 76.05 $\pm$ 4.55 \\

Ours (A3C+A4C → A2C)
& 95.24 $\pm$ 1.40 & 90.94 $\pm$ 2.52
& 95.00 $\pm$ 1.46 & 90.50 $\pm$ 2.61
& 90.09 $\pm$ 4.37 & 82.23 $\pm$ 6.97
& 93.90 $\pm$ 3.13 & 88.65 $\pm$ 5.42 \\

\textbf{Ours}
& 98.35 $\pm$ 0.42 & 96.75 $\pm$ 0.80
& 98.12 $\pm$ 0.69 & 96.31 $\pm$ 1.32
& 94.68 $\pm$ 2.81 & 90.03 $\pm$ 4.86
& 97.89 $\pm$ 0.92 & 95.88 $\pm$ 1.74 \\

\bottomrule
\end{tabular}}
\end{table*}

\begin{figure*}[t]
    \centering
    \includegraphics[width=0.94\textwidth]{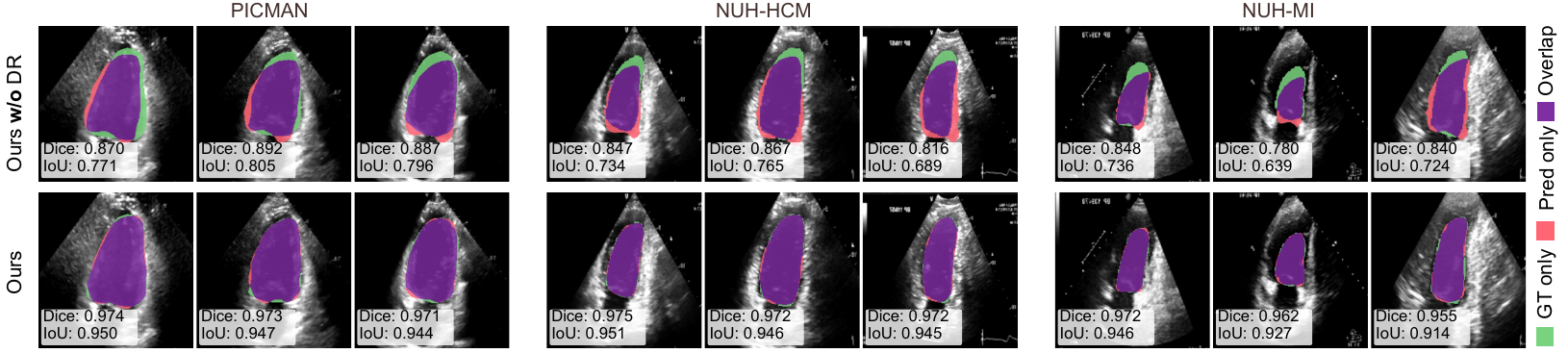}
    \vspace{-10pt}
    \caption{2D mask overlap comparison of cross-disease 4D cardiac reconstruction in the held-out A2C view before and after patient-specific DR. 
    }
\label{fig:cross_cohort_qualitative}
\end{figure*}

\begin{table}[t]
\centering
\caption{Comparison of cross-disease 4D dynamic reconstruction performance. Each cell reports Dice (top) and IoU (bottom) as patient-level mean$\pm$standard deviation (\%).}
\label{tab:4d_dr_rsa_ablation}
\resizebox{0.8\columnwidth}{!}{
\begin{tabular}{lccc}
\toprule
\multicolumn{1}{c}{\multirow[c]{2}{*}[-0.75ex]{\textbf{Method}}} & \textbf{PICMAN} & \textbf{NUH-HCM} & \textbf{NUH-MI} \\
\cmidrule(lr){2-2}\cmidrule(lr){3-3}\cmidrule(lr){4-4}
& \multicolumn{3}{c}{Dice / IoU (\%) $\uparrow$} \\
\midrule
\multirow{2}{*}{\journaladd{Our $w/o$ DR}}
& 86.83 $\pm$ 2.29
& 74.69 $\pm$ 6.23
& 85.60 $\pm$ 4.21 \\
& 77.21 $\pm$ 3.35
& 61.19 $\pm$ 7.11
& 75.47 $\pm$ 5.69 \\
\midrule
\multirow{2}{*}{\textbf{Ours}}
& \textbf{95.20 $\pm$ 1.25}
& \textbf{87.72 $\pm$ 3.73}
& \textbf{93.63 $\pm$ 1.78} \\
& \textbf{91.02 $\pm$ 2.11}
& \textbf{78.83 $\pm$ 5.33}
& \textbf{88.21 $\pm$ 3.06} \\
\bottomrule
\end{tabular}}
\end{table}

\subsection{Ablation Study} \label{sec:exp:ablation_study}

We investigated two key aspects of the proposed model, i.e, DR-driven adaptation to patient-specific geometry and the temporal surface alignment to dynamic mesh consistency.
DR-driven shape adaptation improved agreement with clinical observations beyond pose and scale calibration alone. 
Compared with Echo4DMesh w/o DR, which retained the pretrained decoder and inferred shape code, the full model increased three-view Dice by 10.25\%, 10.72\%, 17.95\%, and 11.83\% on CC, PICMAN, NUH-HCM, and NUH-MI, respectively (Table~\ref{tab:dr_cross_view}). 
The improvement also extended throughout the cardiac cycle: full-cycle Dice reached 95.20\%, 87.72\%, and 93.63\% on PICMAN, NUH-HCM, and NUH-MI, respectively (Table~\ref{tab:4d_dr_rsa_ablation}). 
NUH-HCM showed the largest gains in both evaluations, including a 13.03\% improvement over the full cycle, highlighting the importance of adapting the pretrained geometry to individual anatomy.

Agreement with input masks alone, however, did not establish consistency beyond the views used for shape fitting. 
We therefore evaluated the held-out-view setting (A3C+A4C $\rightarrow$ A2C), in which A2C was used for pose calibration and evaluation but excluded from shape optimization. 
Echo4DMesh achieved A2C Dice scores of 95.24\%, 95.00\%, 90.09\%, and 93.90\% across the four cohorts (Table~\ref{tab:dr_cross_view}). 
The qualitative comparisons in Fig.~\ref{fig:cross_cohort_qualitative} showed fewer reference-only and prediction-only regions after adaptation, including an increase in Dice from 78.0\% to 96.2\% in one example. 
Together, these results indicated that adaptation improved patient-specific fitting while retaining agreement with cross-sections excluded from shape optimization.

\begin{table}[t]
\centering
\caption{Ablation study of RSA and TC-RSA for 4D mesh reconstruction. Results are averaged across the PICMAN, NUH-HCM, and NUH-MI cohorts.}
\label{tab:trirsa_ablation}
\resizebox{0.9\columnwidth}{!}{
\begin{tabular}{lccc}
\toprule
\textbf{Method}
& \textbf{3D Dice (\%) $\uparrow$}
& \textbf{ASSD (\% $L_0$) $\downarrow$}
& \textbf{$\mathbf{V}_{\textbf{gap}}$ (\%) $\downarrow$} \\
\midrule
Ours $w/o$ RSA
& 88.26 $\pm$ 10.06 & 2.09 $\pm$ 1.53 & 9.16 $\pm$ 8.59 \\
Ours $w/$ RSA
& 92.04 $\pm$ 4.98 & 1.10 $\pm$ 0.65 & 6.94 $\pm$ 5.80 \\
+ Bidirectional
& 93.37 $\pm$ 3.49 & 0.92 $\pm$ 0.46 & 6.05 $\pm$ 4.68 \\
+ Adjacent-Frame
& 93.24 $\pm$ 4.88 & 0.93 $\pm$ 0.64 & 5.95 $\pm$ 5.73 \\
+ Repeated RSA
& 93.31 $\pm$ 4.89 & 0.92 $\pm$ 0.64 & 5.88 $\pm$ 5.74 \\
\midrule
\textbf{Ours $w/$ TC-RSA}
& \textbf{94.55 $\pm$ 3.39}
& \textbf{0.76 $\pm$ 0.46}
& \textbf{4.99 $\pm$ 4.49} \\
\bottomrule
\end{tabular}}
\end{table}

\begin{figure}[t]
    \centering
    \includegraphics[width=0.9\columnwidth]{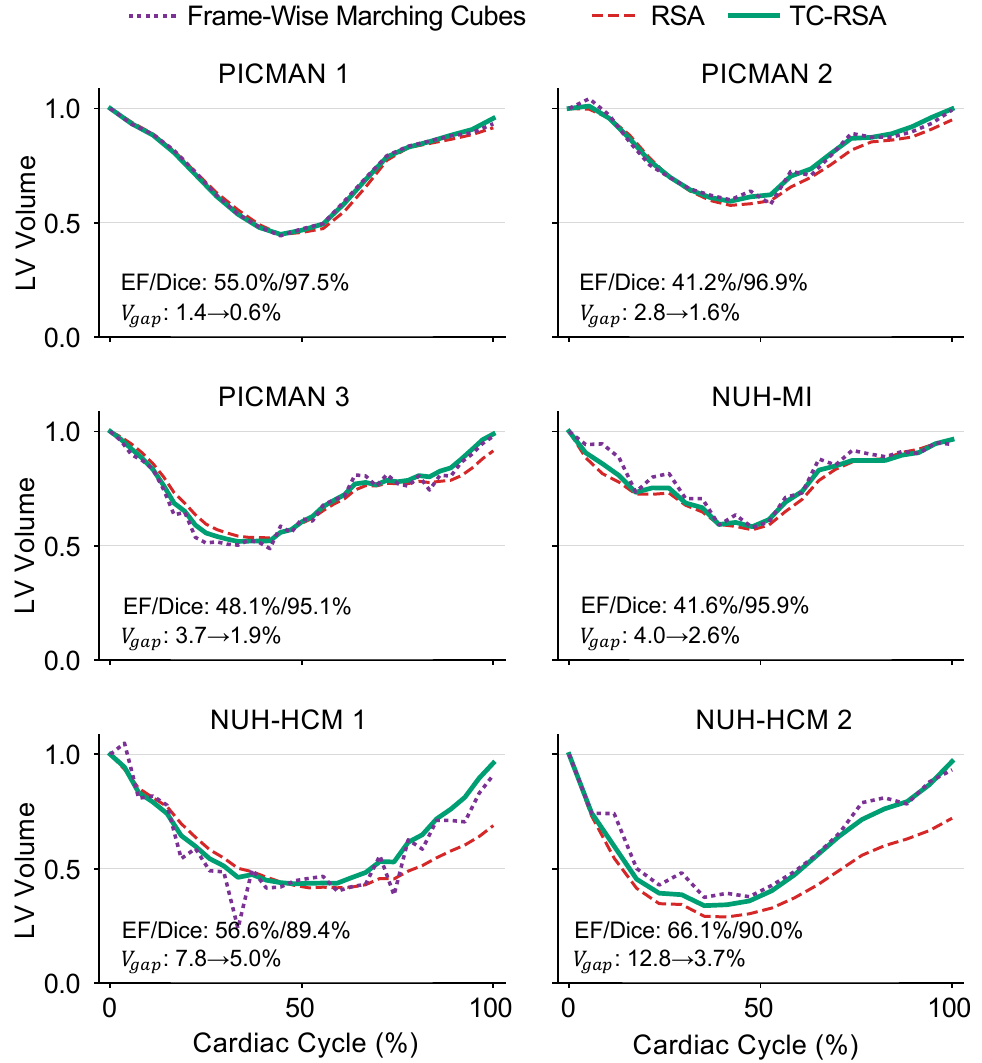}
    \caption{Effect of TC-RSA on LV volume consistency in representative healthy, MI, and HCM cases. Volume curves were normalized by their first-frame values without clipping or smoothing. Ejection-fraction (EF) was computed from TC-RSA meshes, and Dice denoted mean cross-sectional overlap between the adapted SDF and input echo masks across frames. }
    \label{fig:rsa_trirsa_volume_consistency}
\end{figure}

Accurate implicit geometry also required consistent mesh tracking across time. 
RSA improved agreement between propagated meshes and the frame-wise SDF surfaces, increasing 3D Dice from 88.26\% to 92.04\% and reducing Volume Gap from 9.16\% to 6.94\% (Table~\ref{tab:trirsa_ablation}). 
Nevertheless, the volume trajectories in Fig.~\ref{fig:rsa_trirsa_volume_consistency} showed residual discrepancies during late-cycle recovery, particularly in the displayed HCM cases. 
TC-RSA reduced these discrepancies through bidirectional fusion and adjacent-frame blending. 
Each component improved performance over RSA, while their combination achieved the highest Dice of 94.55\%, the lowest ASSD of 0.76\% of $L_0$, and the lowest $V_{gap}$ of 4.99\%. 
Its advantage over repeated RSA correction further supported the benefit of temporal information beyond additional surface-correction steps.

The benefit of temporal refinement was most apparent toward the end of the cycle, where sequential propagation was more susceptible to accumulated errors. TC-RSA followed the frame-wise SDF volume trajectories more closely, reducing the SDF-consistency gap from 12.8\% to 3.7\% in the most pronounced example (Fig.~\ref{fig:rsa_trirsa_volume_consistency}). This improvement coincided with better mesh quality over the final 20\% of the cycle (Table~\ref{tab:late_cycle_mesh_quality}): all three cohorts showed lower $V_{gap}$, fewer low-quality and self-intersecting faces, and higher normal consistency. On NUH-HCM, $V_{gap}$ decreased from 13.36\% to 4.38\%, while self-intersecting faces decreased from 1.034\% to 0.035\%. Thus, temporal refinement improved both global volumetric agreement and local surface quality relative to the fitted implicit geometry.

\begin{table}[!t]
\caption{Late-cycle (80\%-100\%) mesh-to-SDF consistency and mesh quality. Results were averaged within each cohort. Low-Q and Self-int. denoted percentages of faces with triangle quality $Q<0.1$ and self-intersections, respectively. Normal Cons. measured agreement between face normals and SDF gradients.}
\label{tab:late_cycle_mesh_quality}
\centering
\footnotesize
\resizebox{0.9\columnwidth}{!}{
\begin{tabular}{@{}llcccc@{}}
\toprule
\multirow[c]{2}{*}{\textbf{Cohort}}
& \multirow[c]{2}{*}{\textbf{Method}}
& $\mathbf{V}_{\textbf{gap}}$ 
& \textbf{Low-Q}
& \textbf{Self-int.}
& \multirow[c]{2}{*}{\textbf{Normal Cons.} $\uparrow$} \\
& & (\%) $\downarrow$ & (\%) $\downarrow$ & (\%) $\downarrow$ & \\
\midrule
\multirow{2}{*}{PICMAN} & RSA     & 4.09 & 4.9 & 0.202 & 0.990 \\
                        & TC-RSA & \textbf{2.39} & \textbf{4.3} & \textbf{0.026} & \textbf{0.994} \\
\midrule
\multirow{2}{*}{NUH-HCM} & RSA     & 13.36 & 6.1 & 1.034 & 0.911 \\
                         & TC-RSA & \textbf{4.38} & \textbf{4.4} & \textbf{0.035} & \textbf{0.976} \\
\midrule
\multirow{2}{*}{NUH-MI} & RSA     & 4.90 & 4.6 & 0.292 & 0.986 \\
                        & TC-RSA & \textbf{3.51} & \textbf{4.1} & \textbf{0.034} & \textbf{0.993} \\
\bottomrule
\end{tabular}}
\end{table}

\begin{table}[!]
\centering
\caption{Echo-CT external validation on
60 subjects matched to the PICMAN dataset.
For 3D Dice, LV endocardium (Endo) and epicardium (Epi) are evaluated
using the LV cavity and the filled epicardial envelope
(LV$\cup$MYO), respectively.}
\label{tab:ct_external_validation}
{\color{black}
\footnotesize
\resizebox{0.98\columnwidth}{!}{
\begin{tabular}{@{}llccc@{}}
\toprule
\textbf{Structure} & \textbf{Method} &
\textbf{3D Dice (\%) $\uparrow$} &
\textbf{ASSD (mm) $\downarrow$} &
\textbf{HD95 (mm) $\downarrow$} \\
\midrule

\multirow{2}{*}{LV Endo}
& Laumer et al.~\cite{MIA25}
& 84.89 $\pm$ 1.67 & 3.02 $\pm$ 0.90 & 7.77 $\pm$ 1.33 \\
& \textbf{Ours}
& \textbf{90.07 $\pm$ 1.61} & \textbf{1.81 $\pm$ 0.28} & \textbf{5.27 $\pm$ 1.50} \\

\midrule

\multirow{2}{*}{LV Epi}
& Laumer et al.~\cite{MIA25}
& 85.09 $\pm$ 3.58 & 2.99 $\pm$ 0.56 & 7.23 $\pm$ 1.91 \\
& \textbf{Ours}
& \textbf{89.26 $\pm$ 1.67} & \textbf{1.95 $\pm$ 0.39} & \textbf{4.18 $\pm$ 1.10} \\

\bottomrule
\end{tabular}}
}
\end{table}

\begin{figure}[t]
    \centering
\includegraphics[width=\columnwidth]{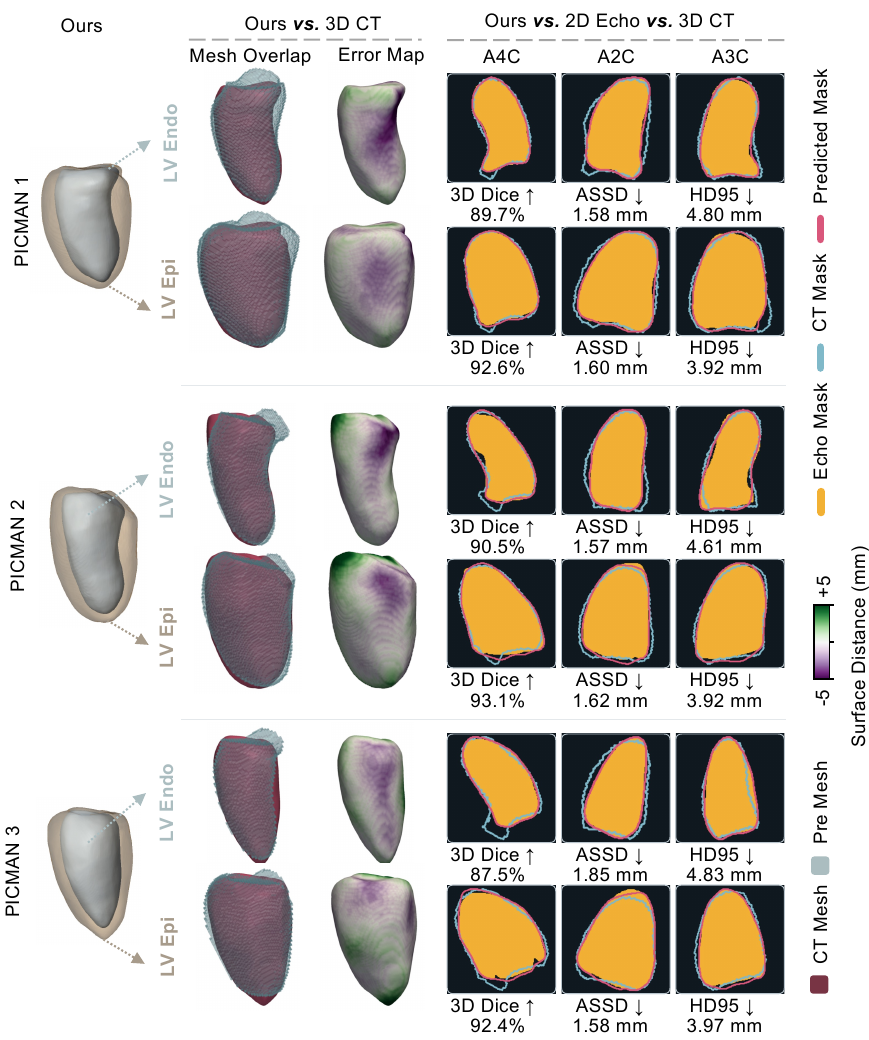}
    \caption{LV endocardial and epicardial reconstruction with patient-matched CT references. Columns show reconstructed surfaces, 3D mesh overlap, signed surface distances (green: outside CT; purple: inside CT), and A4C, A2C, and A3C cross-sections. Scores below each row report Dice, ASSD, and HD95 for the complete 3D target.}
    \label{fig:ct_external_validation}
\end{figure}

\begin{table}[t]
\centering
\caption{Dynamic reconstruction accuracy against MRI-derived meshes on the NUH-MI cohort. Vel. refers to velocity.}
\label{tab:mri_rsa_ablation}
\setlength{\tabcolsep}{3pt}
\resizebox{\columnwidth}{!}{%
\begin{tabular}{lcccc}
\toprule
\textbf{Variant} &
\textbf{EF Error (\%)} $\boldsymbol{\downarrow}$ &
\textbf{Vel. RMSE} $\boldsymbol{\downarrow}$ &
\textbf{3D Dice (\%)} $\boldsymbol{\uparrow}$ &
\textbf{$\mathbf{V}_{\textbf{gap}}$ (\%) $\downarrow$} \\
\midrule
Ours $w/o$ RSA
& 5.52 $\pm$ 3.93
& 51.61 $\pm$ 12.16
& 86.56 $\pm$ 3.51
& 8.36 $\pm$ 4.70 \\

Ours $w/$ RSA
& 2.24 $\pm$ 1.57
& 46.91 $\pm$ 11.01
& 93.88 $\pm$ 1.09
& 2.77 $\pm$ 1.00 \\

\textbf{Ours ($w/$ TC-RSA)}
& \textbf{1.62 $\pm$ 1.30}
& \textbf{34.10 $\pm$ 8.40}
& \textbf{94.43 $\pm$ 0.88}
& \textbf{2.19 $\pm$ 0.63} \\
\bottomrule
\end{tabular}%
}
\end{table}

\begin{figure}[t]
    \centering
    \liu{\includegraphics[width=0.5\textwidth]{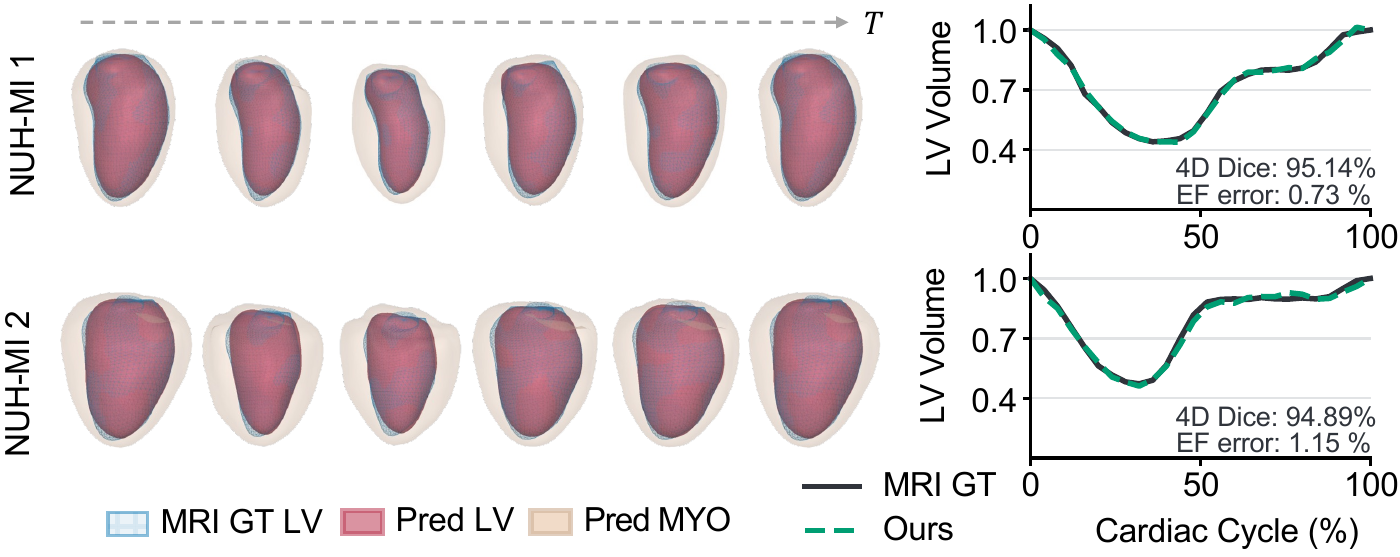}}
    \caption{Cine MRI-based validation of dynamic reconstruction. Mesh overlays and normalized LV volume curves showed agreement between Echo4DMesh reconstructions and MRI references across the cardiac cycle. GT: ground truth.}
    \label{fig:mri_external_validation}
\end{figure}

\subsection{External Validation} \label{sec:exp:external_validation}

\subsubsection{Patient-Matched Echo-CT Validation}

Agreement with input echo masks alone did not establish the accuracy of the reconstructed 3D anatomy. 
We therefore evaluated Echo4DMesh using 60 PICMAN subjects with end-diastolic A2C, A3C, and A4C echo masks and patient-matched CT segmentations.
CT was used only for registration and evaluation, independently of reconstruction and model selection. 
As reported in Table~\ref{tab:ct_external_validation}, Echo4DMesh outperformed Laumer et al.~\cite{MIA25} on all metrics for both left endocardial (Endo) and epicardial (Epi) reconstruction. 
LV Endo Dice increased from 84.89\% to 90.07\%, while ASSD decreased from 3.02 to 1.81~mm and HD95 from 7.77 to 5.27~mm. 
LV Epi Dice increased from 85.09\% to 89.26\%, with ASSD and HD95 decreasing to 1.95 and 4.18~mm, respectively. 
Improvements in both volumetric overlap and surface distances supported more accurate recovery of the LV cavity and outer ventricular geometry beyond the observed 2D cross-sections.

Figure~\ref{fig:ct_external_validation} further examined this agreement through mesh overlays, signed surface distances, and cross-sections along the estimated ultrasound planes. 
The reconstructions followed the overall CT geometry across the three displayed cases, while the distance maps revealed localized inward and outward deviations that were less apparent from the overlays alone. 
Cross-sectional comparisons also distinguished fitting errors from differences between imaging references: reconstructed contours closely followed the input echo masks, but both occasionally deviated from CT. 
For example, the CT contour extended beyond the echo mask and reconstructed contour in the lower A4C region of the third case. 
These findings highlighted the value of an independent 3D reference and identified residual discrepancies potentially associated with sparse-view ambiguity, segmentation differences, and cross-modality registration.

\liu{\subsubsection{Cine MRI-Based Validation with Pseudo-Echo Inputs}}
To assess reconstruction throughout the cardiac cycle, we used pseudo-echo mask sequences generated from 103 NUH-MI cine MRI-derived mesh sequences as reconstruction inputs.
This controlled experiment provided full-cycle anatomical references for evaluating reconstructed geometry, volume evolution, and motion. 
Echo4DMesh with TC-RSA achieved a mean 4D LV Dice of 94.43 $\pm$ 0.88\%, a Volume Gap of 2.19 $\pm$ 0.63\%, and an absolute EF error of 1.62 $\pm$ 1.30\% (Table~\ref{tab:mri_rsa_ablation}). 
Compared with RSA, TC-RSA reduced velocity RMSE from 46.91 to 34.10, alongside improvements in Dice, Volume Gap, and EF error. 
These complementary gains indicated improved motion agreement with the MRI reference while preserving surface accuracy.

Figure~\ref{fig:mri_external_validation} illustrated these results in two representative cases. 
The reconstructed LV surfaces remained closely aligned with the MRI references during contraction and relaxation, while the normalized volume curves reproduced the timing and depth of systolic contraction and subsequent volume recovery. 
The displayed cases achieved 4D Dice scores of 95.14\% and 94.89\%, with EF errors of 0.73\% and 1.15\%, respectively. 
These results supported full-cycle geometric and volumetric accuracy under controlled sparse-view sampling, complementing the anatomical validation using patient-matched echo-CT data.

\begin{table*}[!t]
\centering
{\color{black}
\caption{Comparison of whole-heart reconstruction results on the PICMAN dataset.  }
\label{tab:wholeheart_structure_dice}
\resizebox{0.7\textwidth}{!}{%
\begin{tabular}{lcccccc}
\toprule
\textbf{Method} & \textbf{LV} & \textbf{MYO} & \textbf{RV} & \textbf{LA} & \textbf{RA} & \textbf{Whole Heart} \\
\midrule
Laumer et al.~\cite{MIA25}
& 87.45 $\pm$ 1.06 & 87.82 $\pm$ 0.65 & 79.97 $\pm$ 3.63
& 88.84 $\pm$ 0.90 & 86.93 $\pm$ 1.98 & 86.20 $\pm$ 1.00 \\
Ours $w/o$ DR
& 83.70 $\pm$ 2.27 & 87.36$ \pm$ 1.67 & 72.62 $\pm$ 5.33
& 84.20 $\pm$ 3.23 & 77.72 $\pm$ 4.41 & 81.12 $\pm$ 2.20 \\
\textbf{Ours}
& \textbf{92.48 $\pm$ 1.58} & \textbf{92.74 $\pm$ 1.62} & \textbf{88.30 $\pm$ 3.60}
& \textbf{90.62 $\pm$ 2.74} & \textbf{92.30 $\pm$ 2.91} & \textbf{91.29 $\pm$ 1.49} \\
\bottomrule
\end{tabular}}
}
\vspace{2pt}
\end{table*}

\begin{figure*}[t]
    \centering
    \includegraphics[width=\textwidth]{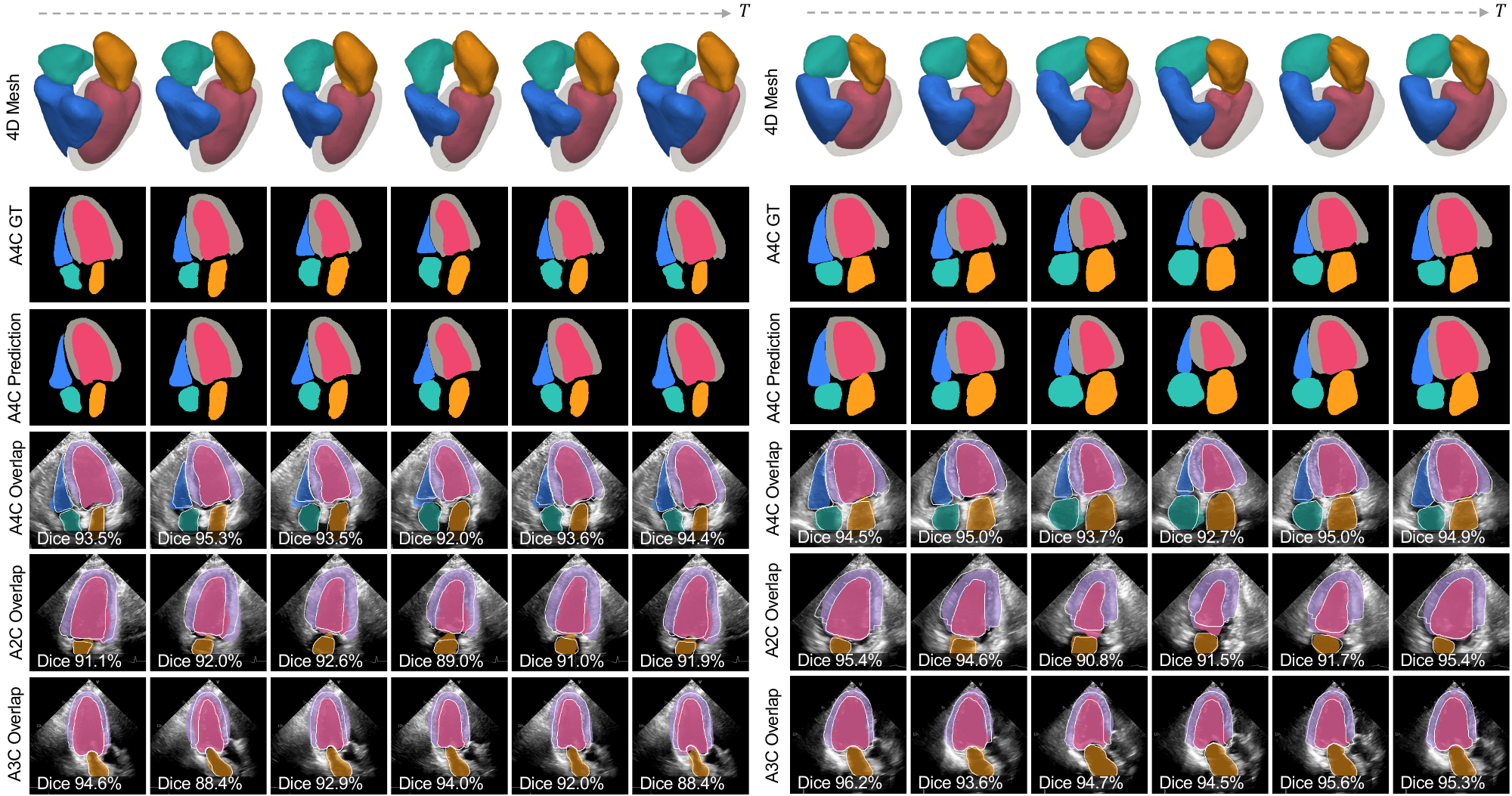}
    \caption{4D whole-heart reconstruction in two representative subjects at a normalized cardiac cycle. Rows show fixed-topology meshes, reference A4C masks, mesh-derived A4C cross-sections, and overlays in A4C, A2C, and A3C. LV, MYO, RV, RA, and LA are colored pink, grey/violet, blue, green, and orange, respectively; white contours indicate manual annotations. Dice scores were averaged over the annotated structures in each view, comparing mesh cross-sections with reference masks.}
\label{fig:wholeheart_v103_qualitative}
\end{figure*}

\subsection{Multi-Structure Whole-Heart Reconstruction} \label{sec:exp:whole_heart}

We extended Echo4DMesh to LV, MYO, RV, LA, and RA using A2C, A3C, A4C, PLAX, and PSAX masks. 
Each structure was initialized from its available annotated views and represented by a continuous SDF within a shared coordinate system. 
Across the whole PICMAN subjects, Echo4DMesh achieved a mean cross-sectional Dice of 91.29 $\pm$ 1.49\%, exceeding Laumer et al.~\cite{MIA25} by 5.09\% and the variant without DR by 10.17\% (Table~\ref{tab:wholeheart_structure_dice}). 
Improvements over Laumer et al.~\cite{MIA25} were observed for every structure, with the largest gain on RV (8.33\%). 
DR-driven adaptation was particularly beneficial for RV and RA, increasing Dice by 15.68 and 14.58\%, respectively, compared with pose calibration alone. 
These results indicated that the extension benefited multiple chambers rather than being driven primarily by LV performance. 
Nevertheless, RV retained the lowest Dice of 88.30\%, identifying it as the most challenging structure in this evaluation.

Figure~\ref{fig:wholeheart_v103_qualitative} further illustrated how the reconstructed structures jointly explained multi-view observations across the cardiac cycle. 
The mesh sequences maintained smooth chamber surfaces and plausible relative positions while representing distinct ventricular and atrial shape changes.
In the second subject, reduced ventricular cross-sections coincided with larger atrial cross-sections during the middle phases, showing that the model accommodated chamber-specific dynamics within the shared coordinate system. 
Intersections of the exported meshes closely reproduced the annotated A4C chamber configurations, with Dice ranging from 92.0\% to 95.3\% and from 92.7\% to 95.0\% in the two subjects. 
Agreement also extended to A2C and A3C, although the lower scores in some phases indicated residual variation across views. 
Together, the quantitative improvements and multi-view overlays supported the feasibility of extending Echo4DMesh to dynamic multi-structure reconstruction; 
independent 3D+t references remained necessary to establish whole-heart anatomical and motion accuracy beyond the observed planes.

\begin{table}[t]
\centering
\caption{Computational efficiency of the proposed method. Runtime includes
test-time optimization and mesh reconstruction.
}
\label{tab:computational_efficiency}
\footnotesize
\resizebox{0.92\columnwidth}{!}{
\begin{tabular}{lcccc}
\toprule
\multicolumn{1}{l}{\multirow{2}{*}{\textbf{Task}}}
& \textbf{Num. of}
& \textbf{Vertices/}
& \textbf{Runtime}
& \textbf{Peak GPU} \\
& \textbf{Frame} & \textbf{Frame}
& \textbf{(s)}
& \textbf{Memory (GiB)} \\
\midrule
Single-LV 3D
& 1 & 9,800 & 33.6 & 0.95 \\
Single-LV 4D
& 33 & 10,054 & 123.4 & 1.13 \\
Whole-Heart 4D
& 33 & 44,234 & 745.3 & 7.00 \\
\bottomrule
\end{tabular}}
\end{table}

\section{Conclusion and Discussion}

In this work, we have developed a framework, Echo4DMesh, for reconstructing patient-specific 4D cardiac meshes from sparse multi-view echo masks without paired clinical 3D supervision. 
The central challenge was to recover geometry beyond the observed imaging planes while maintaining a consistent representation of motion. 
Echo4DMesh addressed these requirements by combining prior-guided implicit shape reconstruction with reference-mesh tracking. 
Experiments across healthy, MI, and HCM cohorts demonstrated adaptation to different anatomies and contraction patterns, while the whole-heart experiments supported extension to multiple cardiac structures.

The results highlighted the complementary roles of anatomical priors and patient-specific observations. 
Sparse cross-sections admitted multiple plausible 3D solutions, making a learned prior useful for constraining unobserved regions. 
However, the improvements over pose calibration alone showed that a population-derived shape estimate did not adequately represent every patient. 
DR-driven adaptation reconciled this estimate with individual observations, with particularly pronounced gains on HCM. 
The held-out-view results further suggested that adaptation improved geometric consistency beyond directly fitting the input cross-sections. 
Nevertheless, this evidence did not uniquely establish the recovered anatomy: the held-out plane still contributed to pose calibration, and regions outside all imaging planes remained dependent on the prior. 
The use of disease-matched SSMs therefore supported adaptation within the evaluated anatomical distributions, while robustness to prior mismatch remained an open question.

A second finding concerned the distinction between reconstructing a sequence of surfaces and reconstructing a mesh sequence with consistent correspondence. Continuous SDFs supported flexible geometric fitting but did not themselves establish vertex correspondence across frames. Conversely, propagating a reference mesh preserved correspondence but introduced discrepancies from the evolving implicit surfaces. RSA and TC-RSA connected these representations through surface correction and temporal refinement. Their improvements in surface agreement, volume trajectories, and mesh quality indicated that correspondence preservation and geometric fidelity could be addressed jointly. However, fixed vertex indexing alone did not establish physiological tissue correspondence, and temporal consistency alone did not guarantee accurate regional deformation. These distinctions were particularly relevant to subsequent strain estimation and biomechanical modeling.

Independent imaging references provided evidence beyond agreement with the input masks or fitted SDFs. Patient-matched echo-CT validation yielded a mean LV endocardial Dice of 90.07\% and an ASSD of 1.81~mm, supporting recovery of patient-specific 3D anatomy from clinical echo observations. The remaining local discrepancies also demonstrated that accurate mask fitting could coexist with cross-modality geometric differences. The MRI-derived pseudo-echo experiment extended evaluation throughout the cardiac cycle, achieving a mean 4D LV Dice of 94.43\% and an absolute EF error of 1.62\%. This experiment supported geometric and volumetric accuracy under controlled sparse-view sampling, but did not reproduce the segmentation uncertainty and acquisition differences of clinical echo. Moreover, agreement in volume and EF did not fully characterize regional motion accuracy. Paired clinical 4D imaging and independent deformation measurements remained necessary for functional validation.

Several practical limitations also remained. The framework depended on segmentation quality and initialization from standard acquisition geometry, while robustness to inconsistent masks and large pose deviations was not systematically established. Whole-heart evaluation demonstrated multi-view agreement but lacked independent 3D+t references for inter-chamber geometry and motion. Test-time optimization also required approximately 30~s for single-LV 3D reconstruction and 120~s for a 33-frame 4D sequence (Table~\ref{tab:computational_efficiency}), limiting suitability for applications requiring immediate reconstruction. Future work will investigate direct conditioning on echo videos, uncertainty estimation in sparsely observed regions, and more efficient patient-specific adaptation. Together with temporally consistent volumetric meshing and further functional validation, these developments could extend Echo4DMesh toward quantitative cardiac assessment and personalized biomechanical simulation.

\section*{Acknowledgment}
This work was supported by Singapore National Medical Research Council Open Fund - Young Individual Research Grant (MOH-001977-01), and Singapore Ministry of Education Tier 1 grant (25-1097-P0001). Y. Liu was partially supported by the China Scholarship Council (Grant No. 202407030035), National Natural Science Foundation of China (Grant No.62162068, 62462066, 62466060), Yunnan Province Ten Thousand Talents Program and Yunling Scholars Special Project (Grant No. YNWR-YLXZ-2018-022), Joint Fund Project for "Double First-Class" Construction of Science and Technology Department of Yunnan Province and Yunnan University (Grant No. 202301BF070001-025), Scientific Research Fund of Yunnan Provincial Department of Education (No.2026Y0185, 2025J0008), supported by No. FWCY-BSPY2024009, No. KC-252513122.

\section*{Declaration of Competing Interest}
The authors have no competing interests to declare that are relevant to the content of this article.

\bibliographystyle{IEEEtran}
\bibliography{A_ref}
\end{document}